\documentclass[acmsmall,screen,nonacm]{acmart}

\usepackage[inline,shortlabels]{enumitem}
\usepackage{subcaption}
\usepackage{url}
\usepackage{multirow}
\usepackage{pifont}
\usepackage{tikz}
\usepackage{threeparttable}
\usepackage[export]{adjustbox}

\newcommand*\eg{\textit{e.g.},~}

\newcommand{\tikzscale}{0.8}

\newcommand{\mh}[2][original]{%
  \ifstrequal{#1}{revised}{%
    \textcolor{green}{[MH: #2]}%
  }{%
    \textcolor{blue}{[MH: #2]}%
  }%
}

\begin{document}

\title{A Survey of Time Series Anomaly Detection Methods in the AIOps Domain}

\author{Zhenyu Zhong}
\email{zyzhong@mail.nankai.edu.cn}
\author{Qiliang Fan}
\email{fanqiliang@mail.nankai.edu.cn}
\author{Jiacheng Zhang}
\email{milocheung@mail.nankai.edu.cn}
\affiliation{%
  \institution{College of Software, Nankai University}
  \streetaddress{Tongyan Road 38}
  \city{Jinnan District}
  \state{Tianjin}
  \country{China}
  \postcode{300350}
}

\author{Minghua Ma}
\email{minghuama@microsoft.com}
\affiliation{%
  \institution{Microsoft Research}
  \streetaddress{Building 2, No. 5 Dan Ling Street}
  \city{Haidian District}
  \state{Beijing}
  \country{China}
  \postcode{100080}
}

\author{Shenglin Zhang}
\email{zhangsl@nankai.edu.cn}
\authornote{Corresponding author of the article.}
\author{Yongqian Sun}
\email{sunyongqian@nankai.edu.cn}
\affiliation{%
  \institution{College of Software, Nankai University}
  \streetaddress{Tongyan Road 38}
  \city{Jinnan District}
  \state{Tianjin}
  \country{China}
  \postcode{300350}
}

\author{Qingwei Lin}
\email{qlin@microsoft.com}
\affiliation{%
  \institution{Microsoft Research}
  \streetaddress{Building 2, No. 5 Dan Ling Street}
  \city{Haidian District}
  \state{Beijing}
  \country{China}
  \postcode{100080}
}

\author{Yuzhi Zhang}
\email{zyz@nankai.edu.cn}
\affiliation{%
  \institution{College of Software, Nankai University}
  \streetaddress{Tongyan Road 38}
  \city{Jinnan District}
  \state{Tianjin}
  \country{China}
  \postcode{300350}
}

\author{Dan Pei}
\email{peidan@tsinghua.edu.cn}
\affiliation{%
  \institution{Department of Computer Science, Tsinghua University}
  \city{Haidian District}
  \state{Beijing}
  \country{China}
  \postcode{100084}
}

\renewcommand{\shortauthors}{Zhong et al.}

\begin{abstract}
  Internet-based services have seen remarkable success, generating vast amounts of monitored key performance indicators (KPIs) as univariate or multivariate time series.
Monitoring and analyzing these time series are crucial for researchers, service operators, and on-call engineers to detect outliers or anomalies indicating service failures or significant events.
Numerous advanced anomaly detection methods have emerged to address availability and performance issues.
This review offers a comprehensive overview of time series anomaly detection in Artificial Intelligence for IT operations (AIOps), which uses AI capabilities to automate and optimize operational workflows.
Additionally, it explores future directions for real-world and next-generation time-series anomaly detection based on recent advancements.

\end{abstract}



\maketitle

\section{Introduction}
Internet-based services like social networking, search engines, and online shopping have experienced tremendous success in recent years.
These services generate a vast amount of monitored key performance indicators (KPIs) in the form of time series data, including system metrics (\eg CPU utilization, memory utilization, network throughput), user-perceived metrics (\eg service response time, error rate), and service performance metrics (\eg page view counts).
The monitoring data of an Internet-based service typically fall into two categories of time series data.
The first is the univariate time series (UTS) of a specific KPI.
The second is the multivariate time series (MTS) of a system entity, which can be a physical machine, microservice, switch, router, docker, disk, or other relevant entity.

Typically, stable Internet-based services exhibit certain \textit{normal patterns} in their monitoring data.
For instance, page view counts are heavily influenced by user schedules and tend to peak during the day while remaining low at night.
Such user behavior leads to seasonal patterns in system metrics that reflect server load.
However, anomalies in the monitoring data can occur due to various reasons, such as data collection issues, software/hardware changes, and service updates.
These anomalies refer to data points that do not conform to the expected normal patterns and often take the form of sudden spikes, unexpected drops, fluctuations, or level shifts in time series.
Anomalies are also known as outliers, discordant observations, discords, exceptions, aberrations, surprises, peculiarities, or contaminants~\cite{DBLP:journals/csur/Blazquez-Garcia21}.
When an anomaly is detected in the collected time series of a system instance, there may be serious failures happening in the upper-level services~\cite{DBLP:conf/conext/ZhangLPCQTZ15, DBLP:journals/tsc/ZhangLPCQTZJF18}.
Service failures may degrade the availability of Internet-based services, impact user experience and even result in economic loss~\cite{DBLP:conf/www/MaXWCZW20, DBLP:journals/tc/SuZSZWZLLTWP22, DBLP:conf/usenix/ZhangLXQ0QDYCCW19}.
Therefore, it is essential to timely identify performance and availability issues (\eg latent bugs, software regressions, service outages) in Internet-based services by detecting anomalies in the monitoring data~\cite{DBLP:conf/www/XuCZLBLLZPFCWQ18, DBLP:conf/ipccc/LiCP18, DBLP:conf/usenix/ZhangLXQ0QDYCCW19, DBLP:conf/issre/MaZPHD18, DBLP:conf/infocom/ChenXLPCQFW19, DBLP:conf/ipccc/BuLZMLZP18, DBLP:conf/www/DaiLLJLXZ21, DBLP:conf/kdd/AbdulaalLL21, DBLP:conf/kdd/LiZHSJWP21, DBLP:journals/tnn/HeCLWYCLZ23}.

With the rapid development of software technologies, modern Internet-based services are undergoing an ever-growing workload and exhibiting an increasing scale and complexity.
The intricate interactions and extensive topology structures of these service architectures can produce enormous, complex, and dynamically changing combinations, making it impossible for operators to manually set anomaly detection rules for each scenario, as they used to do.
As a result, the field of Artificial Intelligence for IT Operations (AIOps) has emerged as a promising solution to address the challenges posed by these dynamic and complex service architectures.
AIOps combines the power of Artificial Intelligence (AI) and Machine Learning (ML) algorithms with traditional IT Operations management practices to automate various tasks, streamline workflows, and improve overall system performance.
In recent years, an increasing number of automatic time series anomaly detection methods in the AIOps domain have been proposed.
However, while the time series anomaly detection problem has been extensively studied in various application domains, such as credit card fraud detection, intrusion detection in cybersecurity, or health monitoring in medical institutions, many of the proposed methods cannot be directly applied to anomaly detection in the AIOps domain.
This is primarily due to the following reasons:
\begin{enumerate}
      \item Time series in the AIOps domain often behave quite differently from those in other application areas.
            The data produced by Internet-based services is vast in scale and typically contains a lot of noise.
            Additionally, the data is frequently imbalanced since anomalies are exceptionally rare in real-world operations.
            Furthermore, the dimensionality of time series can be enormous, sometimes comprising millions of metrics, which adds to the complexity of the anomaly detection task.
      \item Providing detailed anomaly labels manually for such large-scale data is infeasible.
            In many cases, there is no past labeled data for onboarding new time series metrics, which poses a significant challenge in detecting anomalies for these metrics.
      \item In modern Internet-based services, software/hardware may frequently change, altering the normal pattern of the monitoring data.
            Quickly adapting to pattern changes can be challenging without degrading the anomaly detection performance.
      \item In the AIOps domain, various types of anomalies can occur in time series data, making it difficult for anomaly detection approaches to cover as many types of anomalies as possible.
            Additionally, it is often challenging to identify whether an anomalous signal in a time series will impact the service or not, as well as to localize the anomaly and avoid duplication across different pivots such as operations, services, regions, and data centers.
\end{enumerate}
These challenges are particularly significant in the AIOps domain, as the combination of large-scale data, the rarity of anomalies, dynamic pattern changes, the absence of past labeled data, and the absence of well-defined anomaly labels require a more sophisticated and adaptive anomaly detection approach.

Several surveys on methods for detecting anomalies in time series have been presented in the literature~\cite{DBLP:conf/icmla/MunirCDA19, DBLP:journals/csur/Blazquez-Garcia21, DBLP:journals/tnn/GargZSSF22, DBLP:journals/pvldb/SchmidlWP22}.
However, few of these surveys have focused on the AIOps domain, which encompasses both univariate and multivariate time series, and many of the recently proposed methods that have shown promising results have not been included.
The purpose of this review is to provide a structured and comprehensive overview of time series anomaly detection techniques in the AIOps domain, with a focus on the detection algorithms proposed by various authors.
We aim to discuss the various challenges associated with anomaly detection, such as data preprocessing, feature engineering, model selection, and hyperparameter tuning.
Moreover, we also introduce a taxonomy of the different methods and techniques used for time series anomaly detection in the AIOps domain, along with an overview of the datasets and metrics commonly used for evaluation.
Finally, we discuss future directions for research in this area, highlighting the key challenges and opportunities that researchers may encounter when developing new methods.

In this sense, the main contributions of this article can be summarized as follows:
\begin{enumerate}
      \item We provide a comprehensive review that focuses on the problem of time series anomaly detection in the AIOps domain, with a particular emphasis on panel data or cross-sectional data.
            To the best of our knowledge, very few existing surveys exclusively focus on time series anomaly detection in the AIOps domain.
            This review analyzes a vast array of AIOps-related techniques covering 20+ years.
            Some of these methods have not been fully summarized or organized in the literature.
      \item We propose a novel taxonomy for anomaly detection methods in time series data by extracting the most relevant characteristics of the existing methodologies.
            This taxonomy provides a global understanding of anomalies and their detection in time series, which can assist researchers in selecting the most suitable method for a particular problem.
            We also provide details about other features that characterize each technique.
            As far as we know, the existing surveys do not propose a complete taxonomy or extract each method's characteristics.
      \item We explore and discuss some popular datasets/benchmarks and evaluation metrics used in the related anomaly detection work.
            This will help fellow researchers choose the most appropriate dataset/benchmark and evaluation metric.
            We also provide publicly available software links related to the analyzed methods.
            This is an important aspect as it allows easy reproduction of the methods and helps to solve existing problems as soon as possible.
      \item We identify future research directions on time series anomaly detection in the AIOps domain, highlighting areas requiring further exploration and investigation.
\end{enumerate}

The rest of this article is organized as follows.
First, in Section~\ref{subsec:methodology}, we describe the research methodology adopted in this study.
Next, in Section~\ref{sec:problem-formulation}, we introduce the problem of time series anomaly detection in the AIOps domain and summarize the major challenges associated with this problem.
In Sections~\ref{sec:univariate-time-series} and \ref{sec:multivariate-time-series}, we present various techniques for detecting anomalies in univariate and multivariate time series, respectively, along with the rationale for classifying these techniques.
We also provide links to publicly available datasets and software for some of the analyzed anomaly detection techniques.
Section~\ref{sec:evaluation-metrics} explores and discusses popular evaluation metrics used in related methods.
Finally, in Section~\ref{sec:concluding-remarks-and-future-work}, we provide concluding remarks and suggest areas for future research.

\subsection{Methodology}
\label{subsec:methodology}

This study has been oriented and organized to answer the following research questions:
\begin{enumerate}[(RQ1)]
      \item What are the most important characteristics that define each anomaly detection method?
            Based on this, what is the connection between existing techniques?
            How to categorize each technique?
      \item How do existing techniques address the challenges of time series anomaly detection in the AIOps domain?
            What are their main connections/contributions to AIOps?
      \item What open-source datasets are available for time series anomaly detection in the AIOps domain?
            What are their evaluation metrics?
            Are there publicly available software packages?
\end{enumerate}

The methodology used to provide answers to the proposed research questions is an ad hoc methodology that consists of three parts: \textit{Database Selection}, \textit{Survey Search}, and \textit{Literature Search}.

\textit{Database Selection}.
We dug into some scientific research databases to find this review's related articles.
These databases are the following well-known repositories: IEEE Xplore, ACM Digital Library, Scopus, and ScienceDirect.
We also used popular bibliography repositories such as Google Scholar and DBLP, which provide open bibliographic information on major computer science journals and proceedings.

\textit{Survey Search}.
Since our research article is a literature review, we searched for related reviews that have been previously published.
The keywords used are ``AIOps'', ``review'',  ``survey'', ``time series'', ``KPI'', ``anomaly detection'', ``change point detection'', ``concept drift'', and ``outlier detection''.

\textit{Literature Search}.
To search for articles that propose time series anomaly detection methods in the AIOps domain, we mainly used the following keywords between 2000 and 2022: ``AIOps'', ``time series'', ``univariate time series'', ``multivariate time series'', ``KPI'', ``anomaly detection'', ``change point detection'', ``concept drift'', and ``outlier detection''.
Other useful keywords have been ``univariate'', ``multivariate'', and ``metric''.
We also analyzed additional papers that were referenced within papers already identified.
Finally, we manually filtered the results by excluding irrelevant papers that do not aim to solve problems in the AIOps domain.
That is, only methods that have used AIOps-related datasets/benchmarks for evaluation are considered in this review.

\section{Time Series Anomaly Detection in the AIOps Domain}
\label{sec:problem-formulation}
\begin{figure}
    \centering
    \input{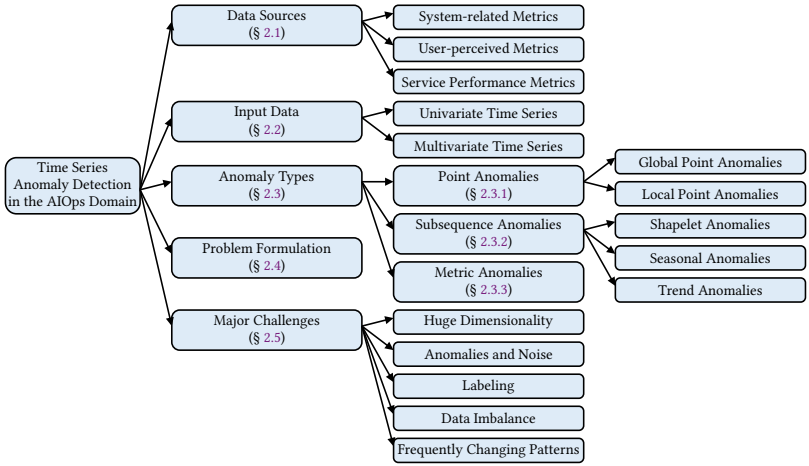}
    \caption{
        The overview of time series anomaly detection problem in the AIOps domain.
    }
    \label{fig:problem-overview}
\end{figure}

In the AIOps domain, the techniques used for time series anomaly detection can vary depending on several factors, including the data source (\S~\ref{subsec:data-sources}), the input data (\S~\ref{subsec:input-data}), the anomaly type (\S~\ref{subsec:anomaly-types}), the ways to address the problem (\S~\ref{subsec:anomaly-detection-problem}), and the primary challenge they aim to solve (\S~\ref{subsec:major-challenges}).
This section presents a comprehensive description that covers all four of these aspects.
Figure~\ref{fig:problem-overview} provides an overview of the problem, and each aspect is discussed in detail below.

\subsection{Data Sources}
\label{subsec:data-sources}

Time series data gathered from Internet-based services can come from various sources, each with its own unique characteristics that can affect the behavior of the collected data.
These Internet service-originated behaviors make these time series data distinct from those in other research fields and render many current time series anomaly detection techniques unsuitable.
In the AIOps domain, three common time series data sources are system-related, user-perceived, and service performance metrics.

\textit{System-related metrics.}
In large-scale Internet-based services like cloud-native platforms, various system instances such as service instances, containers, virtual machines, physical machines, switches, and routers generate time series data known as system-related metrics~\cite{DBLP:conf/www/YuCCGHJWSL21, DBLP:conf/www/MaXWCZW20}.
These metrics can exhibit complex behaviors due to non-Gaussian noise and intricate data distribution, making modeling challenging~\cite{DBLP:conf/infocom/ChenXLPCQFW19}.
System-related metrics are closely monitored with fine granularity to detect micro-congestion caused by burst traffic.
The noise in these time series is often relatively heavy and not diagonal multivariate Gaussian.
Some metrics may exhibit seasonal patterns, while others remain relatively flat, like the \textit{CPU usage} influenced by user schedules or the \textit{disk error rate} that should remain consistently low in a healthy system.

\begin{figure}
    \centering
    \captionsetup[subfigure]{justification=centering}
    \begin{subfigure}[t]{0.2\linewidth}
        \centering
        \includegraphics[width=\linewidth]{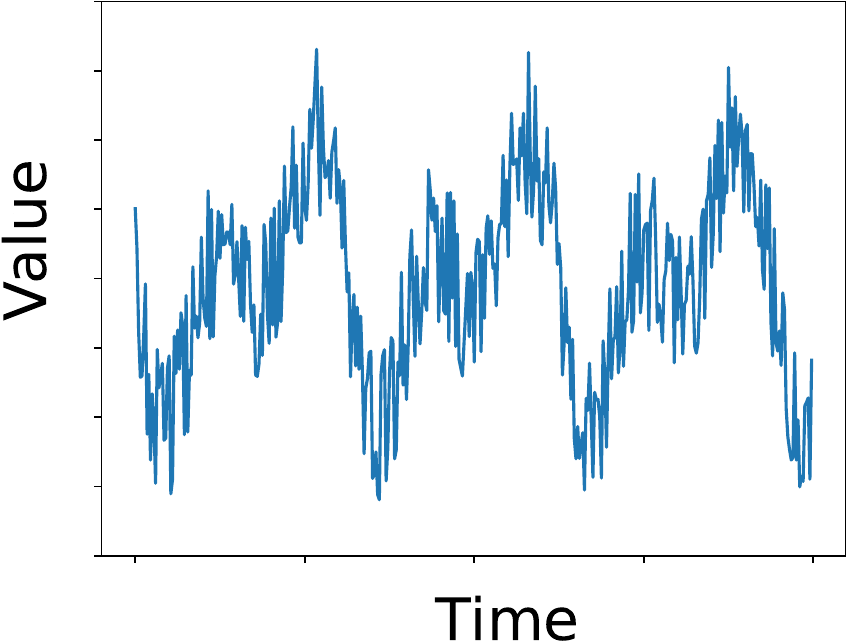}
        \caption{
            System-related UTS.
        }
        \label{subfig:system-related-univariate-time-series}
    \end{subfigure}
    \quad
    \begin{subfigure}[t]{0.2\linewidth}
        \centering
        \includegraphics[width=\linewidth]{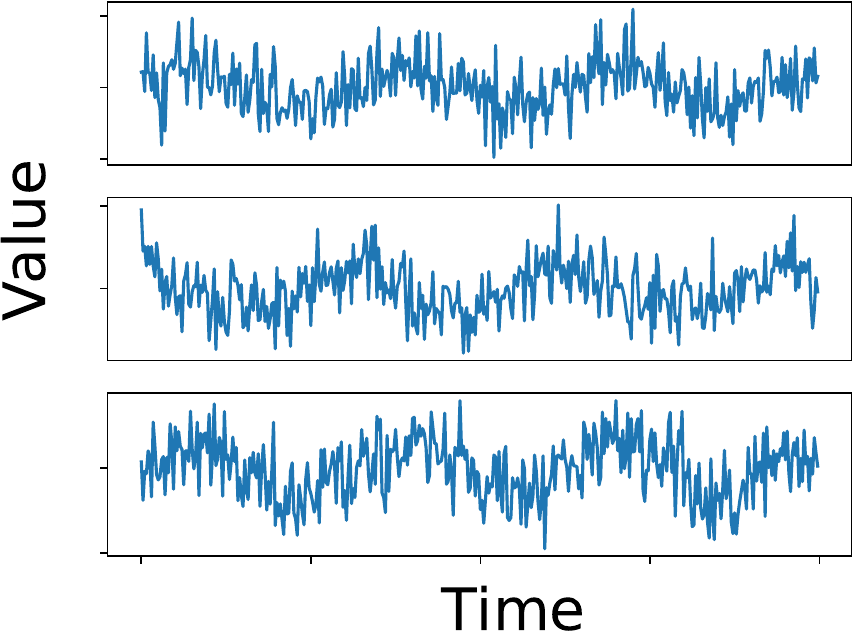}
        \caption{
            System-related MTS.
        }
        \label{subfig:system-related-multivariate-time-series}
    \end{subfigure}
    \quad
    \begin{subfigure}[t]{0.2\linewidth}
        \centering
        \includegraphics[width=\linewidth]{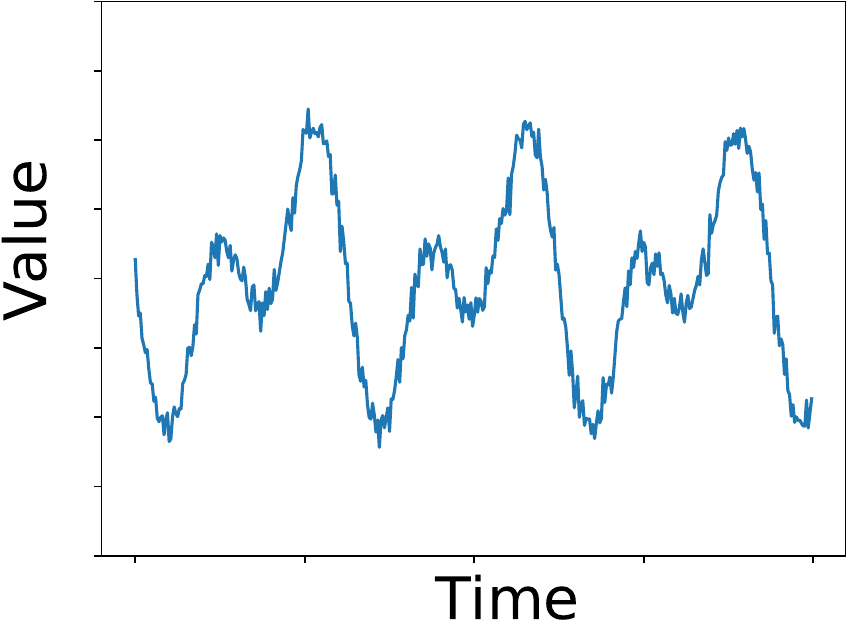}
        \caption{
            User-perceived or service performance UTS.
        }
        \label{subfig:user-perceived-univariate-time-series}
    \end{subfigure}
    \quad
    \begin{subfigure}[t]{0.2\linewidth}
        \centering
        \includegraphics[width=\linewidth]{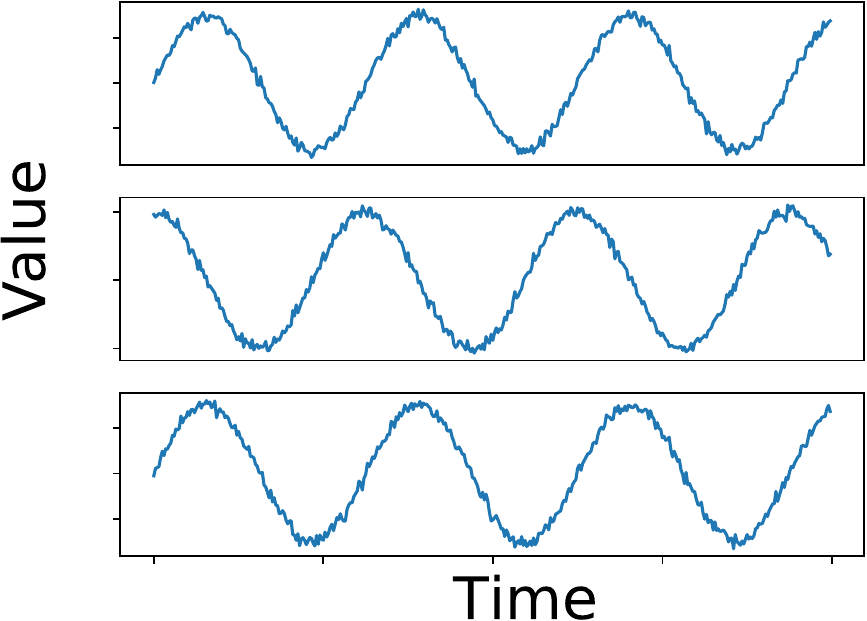}
        \caption{
            User-perceived or service performance MTS.
        }
        \label{subfig:user-perceived-multivariate-time-series}
    \end{subfigure}
    \caption{
        Examples of time series.
    }
\end{figure}

\textit{User-perceived metrics \& service performance metrics.}
Figure~\ref{subfig:user-perceived-univariate-time-series} and Figure~\ref{subfig:user-perceived-multivariate-time-series} depicts examples of user-perceived metrics and service performance metrics.
The user-perceived metrics are metrics that can be perceived by service users, such as \textit{service response time} and \textit{service error rate}, while the service performance metrics encompass metrics that reflect the service's overall performance, such as \textit{page view count}, \textit{number of online users}, and \textit{number of orders}.
Notably, these metrics are subject to the influence of user behavior and scheduling, which causes them to exhibit seasonal patterns that occur at regular intervals, usually on a daily and weekly basis.
Due to the predictable nature of these patterns, it is reasonable to assume that these time series have diagonal multivariate Gaussian noise, thereby providing stability in their behaviors over time.
Monitoring these metrics is crucial to ensure the service meets users' expectations and performs effectively.

\subsection{Input Data}
\label{subsec:input-data}

This aspect pertains to the nature of the input data that time series anomaly detection techniques can employ in the domain of AIOps.
Internet-based services can generate two primary types of time series data: univariate and multivariate.

\textit{Univariate Time Series (UTS).}
A univariate time series is a sequential and ordered collection of real-valued observations, denoted as $\mathbf{X}=\left\{x_{1},x_{2},\ldots,x_{T}\right\}$, with a length of $T\in\mathbb{N}^{+}$.
Each observation in the series is recorded at a specific time $t\in\left\{t:t\in\mathbb{N}^{+}\land t\leq T\right\}$ and with a fixed time interval, such as one second or one minute.
The value of each observation, $x_{t}\in\mathbb{R}$, represents the data point collected at time $t$.
It is commonly assumed that each observation $x_{t}$ is a realization of a random variable $X_{t}$, where the underlying stochastic process generating the series can exhibit different characteristics such as trend, seasonality, or cyclicity.
Illustrative examples of UTS can be observed in Figures~\ref{subfig:system-related-univariate-time-series} and \ref{subfig:user-perceived-univariate-time-series}.

\begin{figure}
    \centering
    \includegraphics[width=0.5\linewidth,frame]{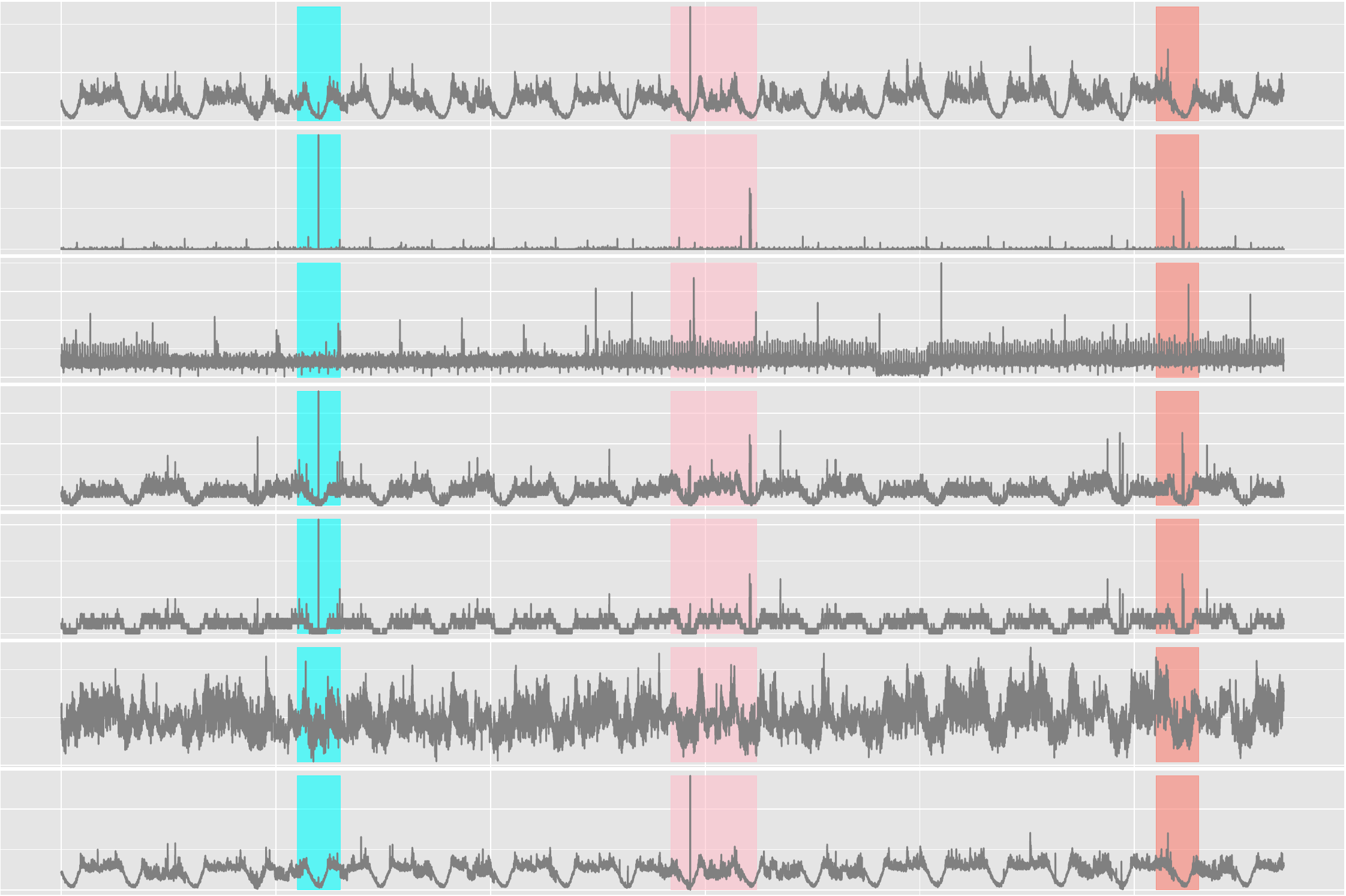}
    \caption{
        The MTS exhibits hidden spatial dependence.
        Vertical color spans indicate anomaly segments.
        There are two correlated metric pairs: Metric 1 and 7, and Metric 4 and 5, respectively.
    }
    \label{fig:spatial-dependence}
\end{figure}

\textit{Multivariate Time Series (MTS).}
A multivariate time series (MTS) is a set of ordered $M$-dimensional vectors corresponding to $M$ metrics.
Each vector is recorded at a specific time interval and consists of $M$ real-valued observations.
In other words, a multivariate time series $\mathbf{X}$ is defined as $\mathbf{X}=\left\{\mathbf{x}_{1},\mathbf{x}_{2},\ldots,\mathbf{x}_{T}\right\},T\in\mathbb{N}^{+}$, where $\mathbf{x}_{t}=\left\{x_{1t},x_{2t},\ldots,x_{Mt}\right\}$ is the observation of the $M$ metrics at time $t$.
Here, $M$ is the total number of monitoring metrics, and $x_{mt}\in\mathbb{R}$ is the observation of the $m$-th metric at time $t$.
Note that for each metric $m\in\left\{m:m\in\mathbb{N}^{+}\land m\leq M\right\}$, $\mathbf{x}_{m}=\left\{x_{m1},x_{m2},\ldots,x_{mT}\right\}$ is a UTS.
An MTS exhibits intra-metric dependence (temporal dependence) and inter-metric dependence (spatial dependence), which means that the random variable $X_{mt}$ depends not only on its past values but also on the other metrics.
Figure~\ref{fig:spatial-dependence} illustrates the spatial dependence hidden in an MTS.
The vertical color spans in the figure denote anomaly segments.
The correlated metrics pose similar behaviors in the normal pattern, and when anomalies emerge, correlated metrics usually act similarly.
Besides the temporal dependence in a single metric, the correlations between the neighbors of each component/instance are also critical for detecting anomalies and localizing the root cause, especially in Internet-based systems such as cloud systems or IoT systems.
The accuracy of MTS anomaly detection can be improved by employing techniques that explicitly model the spatial dependence between components in cloud systems, irrespective of deterministic or stochastic models.

MTS offers a comprehensive view of the monitoring entity, and its anomaly detection has gained significant attention recently~\cite{DBLP:conf/kdd/AudibertMGMZ20, DBLP:conf/kdd/SuZNLSP19, DBLP:conf/kdd/AbdulaalLL21, DBLP:conf/kdd/LiZHSJWP21, DBLP:conf/infocom/SunSZCLPWZLT21, DBLP:conf/usenix/MaZ0XLLNZWP21, DBLP:journals/tc/SuZSZWZLLTWP22}.
However, detecting anomalies in UTS is still crucial in modern operation tasks.
Sometimes, operators are especially interested in the UTS of core key performance indicators (KPIs), and MTS anomaly detection may not fully capture the anomalous behavior of these core KPIs.
When MTS anomaly detection covers dozens of KPIs, it tends to ignore anomalous behaviors of core KPIs or report many alarms that operators do not care about.
Therefore, this review focuses on both univariate and multivariate time series.

\subsection{Anomaly Types}
\label{subsec:anomaly-types}

Similar to general time series, anomalies in AIOps time series can be categorized into three types: point anomalies (\S~\ref{subsubsec:point-anomalies}), subsequence anomalies (\S~\ref{subsubsec:subsequence-anomalies}), and metric anomalies (\S~\ref{subsubsec:metric-anomalies}).

\subsubsection{Point Anomalies}
\label{subsubsec:point-anomalies}

Point-wise anomalies are unexpected incidents that occur at individual time points.
Errors in data acquisition, hardware failures, or unexpected events such as cyber-attacks or natural disasters can cause these anomalies.
Point-wise anomalies usually take the form of spikes or glitches, where a spike refers to an individual point with an extreme value compared to the rest of the points, and a glitch refers to an individual point with a relatively deviated value from its neighboring points.
Point anomalies can be univariate or multivariate, depending on whether they affect one or more metrics.
To further elaborate on this concept, given a UTS or a metric (also takes the form of UTS) from an MTS represented as $\mathbf{X}=\left\{x_{1},x_{2},\ldots,x_{T}\right\}$, two types of anomalies, namely global point anomalies and local point anomalies, can be defined under point-wise behavior with different thresholds $\delta$:
\begin{equation}
    \lvert x_{t}-\hat{x}_{t}\rvert>\delta
\end{equation}
where $\hat{x}_{t}$ is the expected value, which can be the output of a regression model, or simply the global mean value or the mean value of a context window.

\begin{figure}
    \centering
    \begin{subfigure}[t]{0.25\linewidth}
        \centering
        \includegraphics[width=0.96\linewidth]{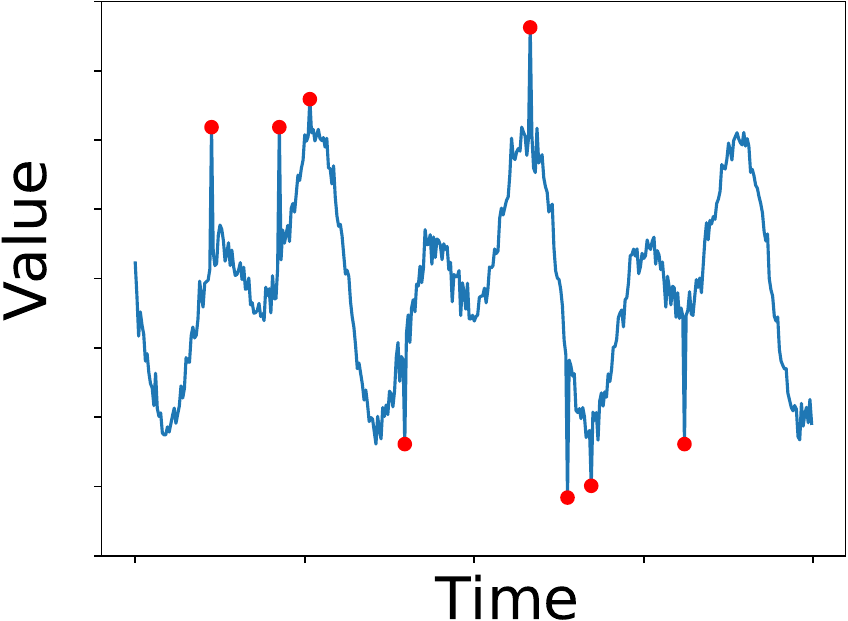}
        \caption{
            Global point anomalies.
        }
        \label{subfig:global-point-anomalies}
    \end{subfigure}
    \quad
    \begin{subfigure}[t]{0.25\linewidth}
        \centering
        \includegraphics[width=0.96\linewidth]{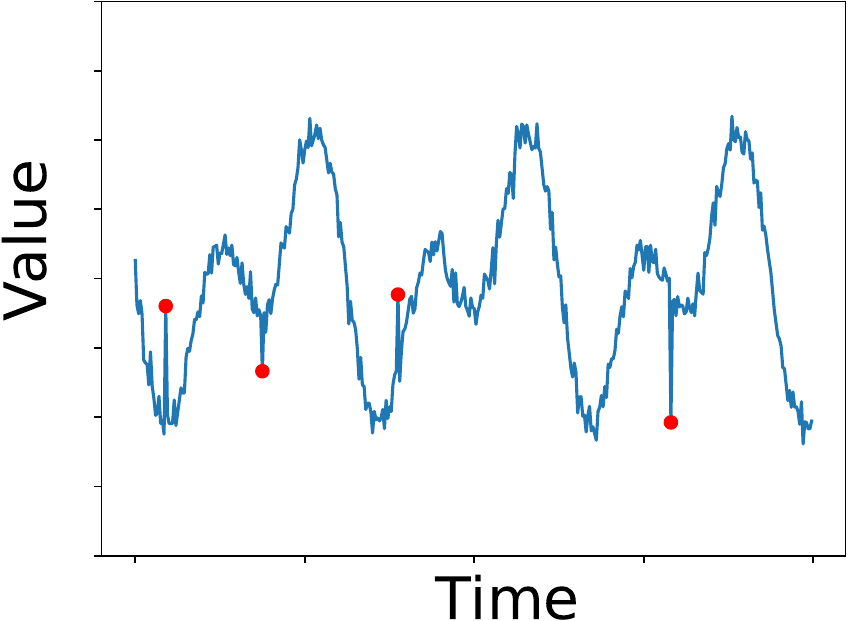}
        \caption{
            Local point anomalies.
        }
        \label{subfig:local-point-anomalies}
    \end{subfigure}
    \caption{
        Point anomalies.
        Red dots denote anomalies.
    }
    \label{fig:point-anomalies}
\end{figure}

Global point anomalies refer to the points that significantly deviate from the rest of the points.
These anomalies are typically large spikes in the time series, and the threshold can be defined as
\begin{equation}
    \delta=\lambda\cdot\sigma\left(\mathbf{X}\right)
\end{equation}
where $\sigma\left(\mathbf{X}\right)$ represents the standard deviation of $\mathbf{X}$ and $\lambda$ is a scaling factor or tolerance.
Figure~\ref{subfig:global-point-anomalies} shows examples of global point anomalies.
Local point anomalies refer to the points that deviate from their corresponding local context or window, which is defined as the neighboring data points within certain ranges.
This type of anomaly is typically a small glitch in sequential data and can be defined as:
\begin{equation}
    \delta=\lambda\cdot\sigma\left(\mathbf{X}_{t-W,t+W}\right)
\end{equation}
where $\mathbf{X}_{t-W,t+W}=\left\{x_{t-W},x_{t-W+1},\ldots,x_{t+W}\right\}$ refers to the local context of $x_{t}$.
The length of the context window is $2W+1$.
Figure~\ref{subfig:local-point-anomalies} shows examples of local point anomalies.

To detect point-wise anomalies, we can first compute the expected value $\hat{x}_{t}$ using various methods.
Then, we can define a threshold $\delta$ and check if the absolute difference between $x_{t}$ and $\hat{x}_{t}$ is less than this threshold. If it is less than the threshold, the point is considered normal; otherwise, it is considered an anomaly.

\subsubsection{Subsequence Anomalies}
\label{subsubsec:subsequence-anomalies}

Subsequence anomalies refer to unusual joint behaviors of consecutive points in time, even though each observation individually may not be a point anomaly.
For instance, a subsequence anomaly can occur when the performance of a system degrades gradually over a period, leading to a drop in its KPIs.
Detection of subsequence anomalies requires techniques that consider the temporal correlation between consecutive observations.

\begin{figure}
    \centering
    \begin{subfigure}[t]{0.24\linewidth}
        \centering
        \includegraphics[width=\linewidth]{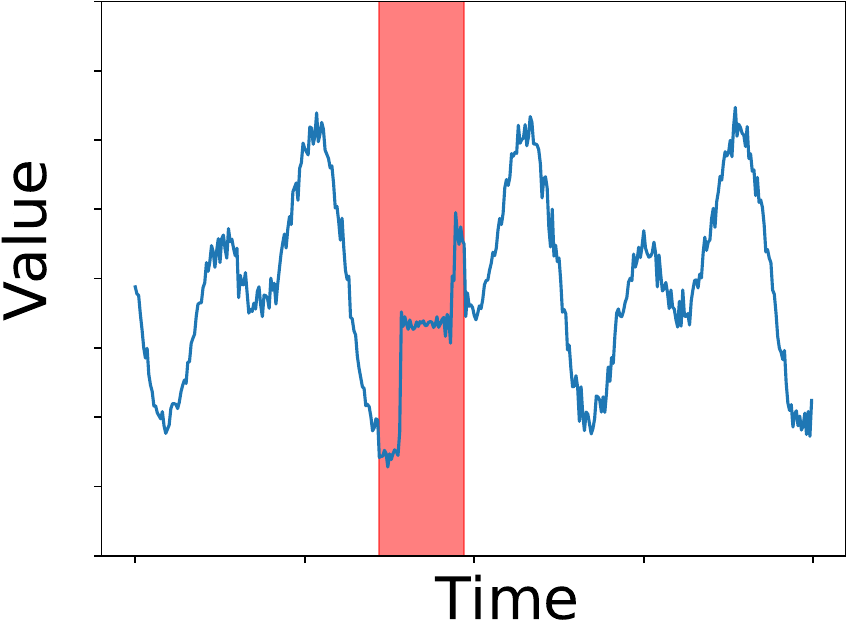}
        \caption{Shapelet anomalies.}
        \label{subfig:shapelet-anomalies}
    \end{subfigure}
    \quad
    \begin{subfigure}[t]{0.24\linewidth}
        \centering
        \includegraphics[width=\linewidth]{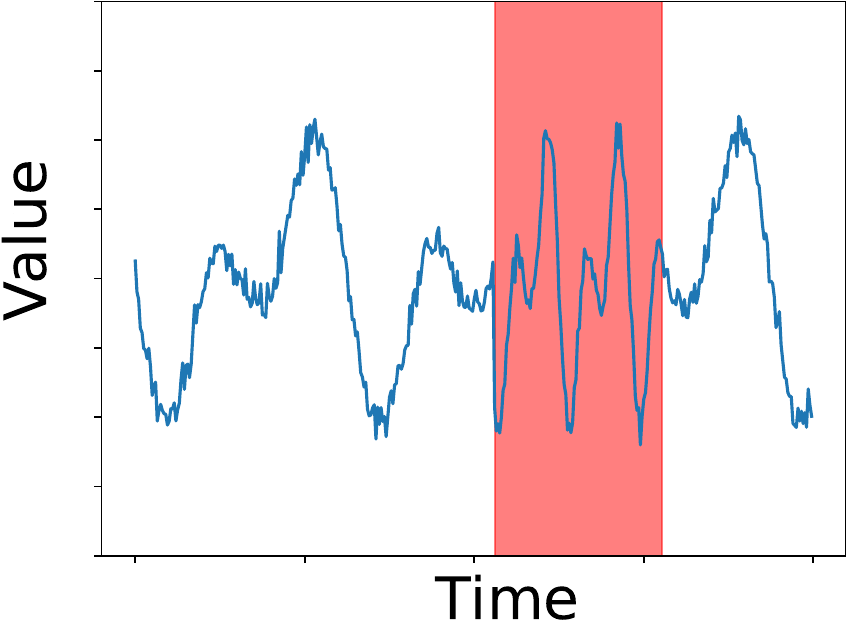}
        \caption{Seasonal anomalies.}
        \label{subfig:seasonal-anomalies}
    \end{subfigure}
    \quad
    \begin{subfigure}[t]{0.24\linewidth}
        \centering
        \includegraphics[width=\linewidth]{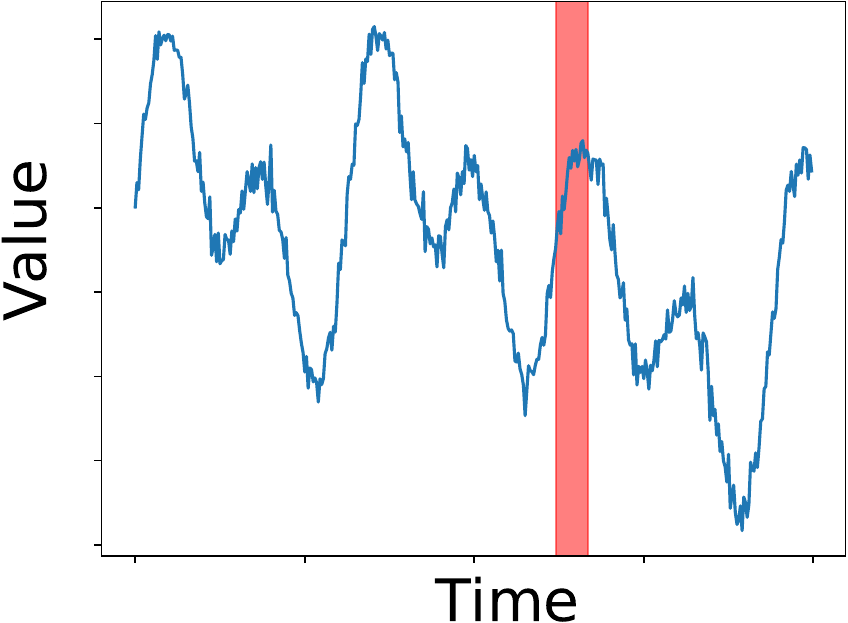}
        \caption{Trend anomalies.}
        \label{subfig:trend-anomalies}
    \end{subfigure}
    \caption{
        Subsequence anomalies.
        Red spans denote anomalies.
    }
    \label{fig:subsequence-anomalies}
\end{figure}

To define a time series with periodic patterns, we can use the following equation:
\begin{equation}
    \mathbf{X}=\phi\left(\omega\mathbf{T}\right)+\tau\left(\mathbf{T}\right)+\epsilon
\end{equation}
where $\mathbf{T}$ represents the timestamps, $\phi$ denotes the underlying shape (or shapelet), $\omega$ refers to the seasonal component, $\tau$ corresponds to the trend component, and $\epsilon$ signifies the residual component.
By decomposing the time series into these components, we can better understand the nature of subsequence anomalies present in the data.
These anomalies can be classified into three types: shapelet anomalies, seasonal anomalies, and trend anomalies.
Shapelet anomalies (Figure~\ref{subfig:shapelet-anomalies}) are time series segments whose underlying shape is significantly different from previous periods at the same time.
Seasonal anomalies (Figure~\ref{subfig:seasonal-anomalies}) are characterized by noticeable changes in the seasonality of the time series data, where patterns occur more or less often than in normal situations.
Trend anomalies (Figure~\ref{subfig:trend-anomalies}) represent level shifts in time series data, where the data values are shifted higher or lower.

Subsequence anomalies can be either global or local and can affect one (UTS anomaly) or more (MTS anomaly) time series/metrics.
It is important to note that MTS anomalies do not necessarily affect all the metrics, and multiple types of subsequence anomalies can exist simultaneously in the same time series.

\subsubsection{Metric Anomalies}
\label{subsubsec:metric-anomalies}

\begin{figure}
    \centering
    \includegraphics[width=0.24\linewidth]{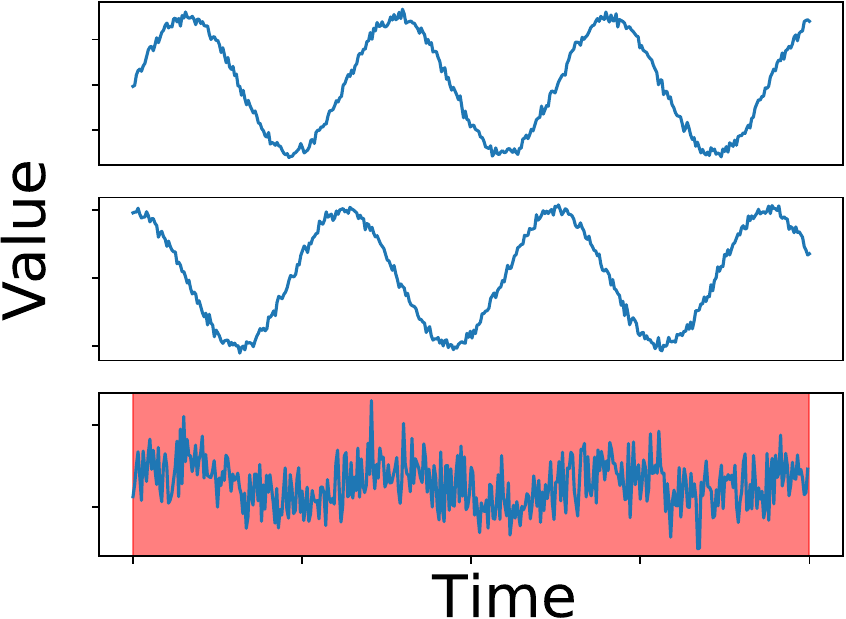}
    \caption{
        Metric anomalies.
        Red spans denote anomalies.
    }
    \label{fig:metric-anomalies}
\end{figure}

An entire time series can also be an anomaly, but this can only be detected when the input data is an MTS.
Figure~\ref{fig:metric-anomalies} provides an example of an anomaly time series, where Metric 3, in an MTS comprising three metrics, is considered anomalous because its behavior significantly deviates from the other metrics and the normal historical observations.

\subsection{Problem Formulation}
\label{subsec:anomaly-detection-problem}

In the domain of AIOps, the problem of anomaly detection for both univariate time series (UTS) and multivariate time series (MTS) can be formulated as follows:

\textit{Univariate time series anomaly detection in the AIOps domain.}
For each time step $t$, given a real-valued observation $x_{t}$ and a historical observation window $\mathbf{X}=\left\{x_{i},x_{i+1},\ldots,x_{i+W-1}\right\}$ of length $W$, the task is to determine whether an anomaly has occurred at time $t$ (indicated by $\gamma_{t}=1$).

\textit{Multivariate time series anomaly detection in the AIOps domain.}
For each time step $t$, given an observation vector $\mathbf{x}_{t}$ (composed of $M$ real-valued metric observations) and a historical observation window $\mathbf{X}=\left\{\mathbf{x}_{i},\mathbf{x}_{i+1},\ldots,\mathbf{x}_{i+W-1}\right\}$ of length $W$, the task is to determine whether an anomaly has occurred at time $t$.

Anomaly detection methods typically provide two ways to give the detection results.
The first type is quite straightforward: they directly determine whether a data point is anomalous or not.
The second type, however, is more complex.
It involves the provision of a real-valued score called the ``anomaly score'', such as $P\left(\gamma_{t}=1\mid x_{i:i+W-1}\right)$.
In this case, an anomaly score threshold needs to be selected to determine whether to trigger further alerts.
Methods that use the second strategy can describe ``how anomalous'' they consider a data point to be, enabling operators to deploy more sophisticated alerting rules and facilitate further analysis.

\subsection{Major Challenges}
\label{subsec:major-challenges}

Detecting anomalies in time series data is a challenging task in the AIOps domain.
The nature of time series data produced by Internet-based services makes this task particularly intricate.
This complexity can be attributed to the following five challenges:

\begin{enumerate}
    \item \textit{The dimensionality of time series is huge.}
          The dimensionality of time series data in large-scale Internet-based services is vast due to three main factors.
          First, the scale of these services has rapidly increased, resulting in millions of monitoring entities.
          Second, each entity is monitored at a high frequency, generating a large number of values for each metric.
          Third, monitoring entities usually have numerous metrics, reflecting various aspects of their health.
          Anomaly detection algorithms typically require learning the distributions of all this data, which consumes significant computational resources and hinders prompt detection.
    \item \textit{The interference of noise and anomalies.}
          Many anomaly detection methods detect anomalies in time series based on their normal patterns.
          However, there is a lot of noise and anomalies in the time series in the AIOps domain, which can significantly degrade the performance.
          Therefore, effective and robust approaches need to be applied to improve computational efficiency while avoiding the interference of noise and anomalies.
    \item \textit{The excessive overhead of labeling work.}
          Supervised anomaly detection methods require accurate anomaly labels, but obtaining high-quality ground truth is challenging due to several reasons.
          Firstly, labeling anomalies in time series requires domain knowledge of IT operations, leading to a scarcity of reliable labelers in the AIOps domain compared to more popular areas like image and speech recognition.
          Additionally, even experienced operators may struggle with confidently defining anomalies in different time series and applications.
          Thirdly, operators need to label a large number of time series for training supervised anomaly detection algorithms.
          The labor-intensive process of carefully examining lengthy time series to label anomalies consistently adds to the challenge~\cite{DBLP:conf/infocom/ZhaoZLLZP19}.
          Therefore, in a production environment, methods that reduce labeling overhead while maintaining efficiency and accuracy are preferable.
    \item \textit{The data imbalance.}
          Detecting anomalies in time series data is commonly treated as a binary classification problem, aiming to distinguish between normal and anomalous data points.
          However, in practical AIOps production environments, the number of anomalous data points is often much lower than the number of normal ones, leading to an imbalanced class distribution problem.
          This can create significant challenges, particularly for supervised methods.
    \item \textit{Frequently changing data patterns.}
          Internet-based services frequently undergo software and hardware upgrades and configuration changes, happening thousands of times daily~\cite{DBLP:conf/issre/MaZPHD18}.
          These changes lead to pattern changes in time series.
          Notably, these pattern changes are expected and should not be treated as anomalies.
          However, the anomaly detection system's existing parameters and model weights are trained on the old time series before the change, causing many false alarms due to significant data distribution shifts.
          Manually tuning the anomaly detectors' parameters for numerous expected pattern changes is impractical.
          Hence, an efficient approach to automatically adapt to time series changes is crucial to reduce false alarms and minimize human intervention.
\end{enumerate}

\section{Univariate Time Series}
\label{sec:univariate-time-series}
\begin{figure}
    \centering
    \input{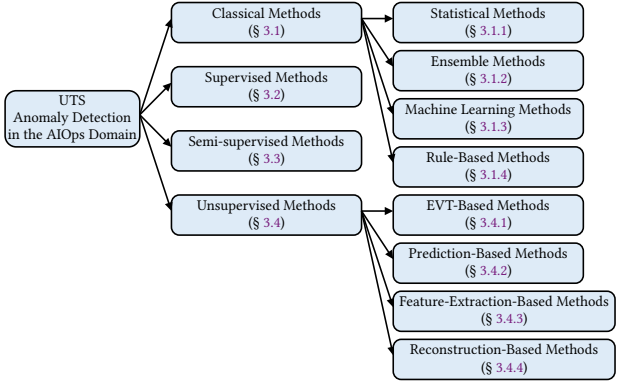}
    \caption{
        Types of methods for detecting anomalies in UTS.
    }
    \label{fig:utad-overview}
\end{figure}

In the AIOps domain, anomaly detection techniques for univariate time series (UTS) can be broadly classified into four categories: classical methods (\S~\ref{subsec:utad-classical-methods}), supervised methods (\S~\ref{subsec:utad-supervised-methods}), semi-supervised methods (\S~\ref{subsec:utad-semi-supervised-methods}), and unsupervised methods (\S~\ref{subsec:utad-unsupervised-methods}).
This section is dedicated to providing a brief overview of these methods, which are commonly used in UTS anomaly detection.
The organization of this section is depicted in Figure~\ref{fig:utad-overview}.
Furthermore, we will also present public code repositories and provide a summary of these methods in \S~\ref{subsec:utad-summary}.
In addition, we have compiled a comprehensive list of public datasets that are commonly used by these approaches.
This list is presented in Table~\ref{tab:utad-datasets}, and the datasets are arranged in descending order of usage frequency.
Links to access these datasets have also been provided.

\subsection{Classical Methods}
\label{subsec:utad-classical-methods}

While numerous time series anomaly detection methods that employ sophisticated deep learning techniques have been proposed recently, traditional techniques still play a crucial role in this field.
This section provides a review of some typical classical methods, which can be classified into four categories: statistical methods (\S~\ref{subsubsec:utad-statistical-methods}), ensemble methods (\S~\ref{subsubsec:utad-ensemble-methods}), machine learning methods (\S~\ref{subsubsec:utad-machine-learning-methods}), and rule-based methods (\S~\ref{subsubsec:utad-rule-based-methods}).
These methods have significantly contributed to the development of time series anomaly detection, and most utilize traditional statistical models.
Their relevance continues to influence the development of time series anomaly detection techniques.

\subsubsection{Statistical Methods}
\label{subsubsec:utad-statistical-methods}

One simple yet effective way of detecting anomalies in UTS is by using average-based methods.
The basic idea behind average-based methods is to calculate the average of the time series and then compare the value of each data point with the average.
Therefore, average-based methods are particularly useful for detecting point anomalies.
Average-based methods can be formulated by:
\begin{equation}
    s_{t}=\left|x_{t}-\bar{x}\right|
\end{equation}
where $s_{t}$ represents the anomaly score for time $t$, $x_{t}$ is the observed value at time $t$, and $\bar{x}$ is the averaged value.

\begin{figure}
    \centering
    \input{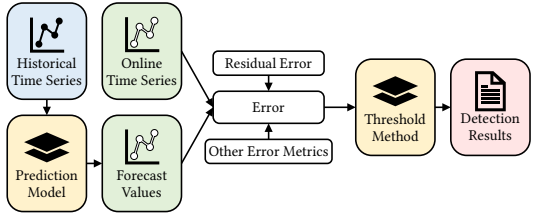}
    \caption{
        The general framework of prediction-based methods.
    }
    \label{fig:utad-prediction-based-methods}
\end{figure}

Most average-based methods are also prediction-based, and Figure~\ref{fig:utad-prediction-based-methods} provides a general framework for such methods.
One popular averaging method is the historical average.
The historical average assumes that the data follow Gaussian distributions and employs the data points' standard deviation as the anomaly score threshold.
\citet{DBLP:conf/infocom/LeePHPLYGYK12} adopted the historical average approach and developed a practical algorithm to identify spatial-temporal clusters with similar threshold behaviors, thereby minimizing the number of thresholds required while maintaining detection accuracy.

Another simple averaging method is the moving average (MA).
MA calculates the average of the time series in a sliding window.
This method can assist operators in smoothing data and identifying the underlying long-term trends.
\citet{DBLP:conf/sigcomm/ChoffnesBG10} employed naive MA for detecting unusual network events at the network edge (local detection) and used the likelihood ratio to determine whether these local anomalies were indeed caused by network problems (global detection).

The autoregressive integrated moving average (ARIMA) model~\cite{box1976time} improves upon naive MA and is particularly suitable for forecasting time series.
ARIMA combines three primary methods, namely autoregression, differencing, and moving average.
\citet{DBLP:conf/imc/ZhangGGR05} separated the inference and anomaly detection steps and tried a variety of models for time series anomaly detection.
Their experiments showed that algorithms that combined ARIMA with $L^{1}$-norm minimization were effective in dealing with time series with noise and missing values.

The exponentially weighted moving average (EWMA) model~\cite{DBLP:journals/technometrics/Roberts00} is similar to MA but is designed to assign lower weights to older observations, with the weights decreasing exponentially.
\citet{DBLP:conf/imc/KrishnamurthySZC03} adopted the sketch model, a probabilistic summary technique proposed for analyzing massive streaming datasets, for dimension reduction.
They then applied a set of time series prediction models, including EWMA, to the sketch output.
Furthermore, \citet{DBLP:conf/imc/KrishnamurthySZC03} introduced a multi-pass grid search technique to find suitable parameters for EWMA, while \citet{DBLP:conf/icc/HimuraFCE09} determined optimal parameters by maximizing the ratio of detected anomalies.

Other classical detectors do not involve averaging.
Several effective techniques besides averaging can be employed to extract relevant information from time series.
One such method is Holt-Winters~\cite{brockwell2009time}, which is an exponential smoothing algorithm that considers three important aspects of time series: baseline, trend, and seasonality.
Another popular technique is Argus~\cite{DBLP:conf/infocom/YanFGGMPSY12}, which utilizes a tailored additive Holt-Winters algorithm as an event detector.
This algorithm is particularly useful when spatial and temporal aggregation is required, owing to its low computational and memory requirements.

Subspace algorithms based on singular value decomposition (SVD), on the other hand, are more suitable for processing sub-time-series, compared to continuous time series models such as ARIMA.
PRISM~\cite{DBLP:conf/conext/MahimkarGWYZEHS11} proposed a new multiscale robust local subspace algorithm that utilizes SVD to improve its vulnerability to subspace contamination.
This approach has been shown to be effective in identifying and isolating anomalies in time series.

Another widely used technique for analyzing time series is the wavelet transform.
It is a localized analysis method that allows us to analyze the time (or space) and frequency characteristics of a signal.
\citet{DBLP:conf/imc/BarfordKPR02} used a series of wavelet filters to isolate specific features of the data and then combined them to detect anomalies automatically.
This approach has been shown to be highly effective in identifying subtle and complex anomalies in time series.

\subsubsection{Ensemble Methods}
\label{subsubsec:utad-ensemble-methods}

This paragraph explores the topic of classical ensemble methods, which involve the creation of multiple detectors that are then combined to produce better results.
MAWILab~\cite{DBLP:conf/conext/FontugneBAF10} introduced a graph-based approach that could combine the outputs of multiple detectors, even if their granularity varied.
The method utilized majority voting to determine the final output and tested four combining strategies, namely, average, minimum, maximum, and an SVD-based dimensionality reduction approach that was found to be the most effective in experiments.

Exploiting the performance improvement of processors, PAD~\cite{DBLP:journals/network/ShanbhagW09} proposed a parallel anomaly detection system that applied multiple existing detectors on different time series subsets.
Compared to traditional methods, this system achieved higher accuracy and covered more anomaly categories.
The method employed a simple normalization scheme to map the output of each algorithm to the range $\left[0,1\right]$ and used the average of the maximum and mean values as the aggregation function.
Moreover, \citet{DBLP:conf/icc/AshfaqJKR10} compared three existing combining schemes, including voting-based, ENCORE~\cite{DBLP:conf/das/RahmanAF02}, and Bayesian network-based combining and proposed a new combining scheme, the standard deviation normalized entropy of accuracy (SDnEA), which employed the mean and variance of the accuracy of each detector obtained from the initial supervised learning phase to calculate the weight of each detector.
The weight formula is expressed as follows:
\begin{equation}
    w_{d_{i}}=\frac{1+p_{d_{i}} \log_{2}\left(p_{d_{i}}\right)+\left(1-p_{d_{i}}\right) \log _{2}\left(1-p_{d_{i}}\right)}{\sigma_{d_{i}}}
\end{equation}
Here, $d_{i}$ denotes detector $i$, $w$ denotes the weight, $p$ denotes the mean accuracy, and $\sigma$ is the standard deviation of accuracy.

\subsubsection{Machine Learning Methods}
\label{subsubsec:utad-machine-learning-methods}

Besides the traditional statistical methods and ensemble methods discussed earlier, there are also methods that reflect the trend of machine learning applications in anomaly detection.
A survey that has had a significant impact on the field of anomaly detection is presented by \citet{DBLP:journals/csur/ChandolaBK09}.
The survey provides a structured and comprehensive overview of anomaly detection and details six categories of anomaly detection methods.
It is worth noting that the survey presents several machine learning-based algorithms, such as neural networks, Bayesian networks, and SVMs. We are not going to discuss these methods in detail, as they have been already covered in \cite{DBLP:journals/csur/ChandolaBK09}  comprehensively.

In addition, a method called the additive Hilbert-Schmidt Independence Criterion (aHSIC)~\cite{DBLP:conf/ijcai/YamadaKNS13} calculates the weighted sum of the HSIC scores between features and their corresponding binary labels.
It is suitable for solving the change point detection problem in high-dimensional time series.
To obtain optimal weights, aHSIC employs a machine learning technique called the dual augmented Lagrangian (DAL), which makes the model more robust to noise.

\subsubsection{Rule-Based Methods}
\label{subsubsec:utad-rule-based-methods}

This paragraph describes different methods that utilize pre-defined rules for detecting anomalies in UTS.
The first method, presented by \citet{DBLP:conf/sigcomm/ChenMSZ13}, focuses on detecting anomalies in web search response time (SRT) by decomposing SRT into long-term trends, periodic behavior, and noise.
The noise is assumed to follow a Gaussian distribution, and the central limit theorem is used to determine the threshold.
SRT is divided into 14 granular measures and three classes to facilitate diagnosis.
Another method, MERCURY~\cite{DBLP:conf/sigcomm/MahimkarSGSWYZE10}, detects changes in time series caused by network upgrades.
It uses a rank-based statistical change point detection method called cumulative sum chart (CUSUM), which selects candidate change points using cumulative sums, determines significant changes using significance testing, and repeats the process for the remaining subsequences.
The cumulative sums are computed as:
\begin{equation}
    S_{i}=S_{i-1}+\left(r_{i}-\bar{r}\right)
\end{equation}
where $S_{0}=0$, $r_{i}$ is the rank of sample $i$ when the data is sorted in ascending order, and $\bar{r}$ is the mean across all ranks.

To overcome the limitations of pre-defined rules, \citet{DBLP:conf/imc/SouleST05} combines filtering and statistical learning techniques for anomaly detection.
The method uses the Kalman filter for prediction and estimation and then detects anomalies by comparing the difference between the two.

\subsection{Supervised Methods}
\label{subsec:utad-supervised-methods}

Despite the impressive results achieved by classical anomaly detectors in experimental settings, their real-world performance leaves much to be desired due to the need for manual parameter tuning and anomaly threshold setting.
This section focuses on supervised learning methods that are specifically designed to detect anomalies for UTS data in the AIOps domain.
These approaches typically require a large amount of labeled data to train the model, but they can achieve high accuracy and high efficiency in scenarios where sufficient labeled data is available.

\begin{figure}
    \centering
    \input{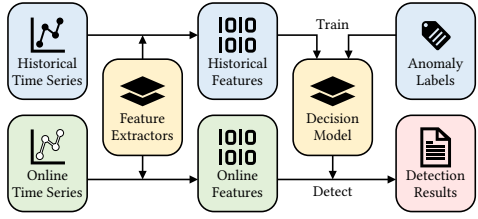}
    \caption{
        The general framework of feature-extraction-based methods.
    }
    \label{fig:utad-feature-extraction-based-methods}
\end{figure}

Opprentice~\cite{DBLP:conf/imc/LiuZXSPLJF15} proposes a novel approach where the output of multiple basic anomaly detectors is treated as features of the original data.
Supervised machine learning models are then trained to predict the labels of the data based on these features.
Approaches like this are referred to as feature-extraction-based methods.
Figure~\ref{fig:utad-feature-extraction-based-methods} illustrates the process.
Anomalies are labeled in historical data using a labeling tool.
Basic anomaly detectors quantify anomaly levels and generate features.
These features, along with labels, form the training set.
A classifier, like a random forest model, is trained on this set with an adjusted threshold for accuracy preference.
Supervised methods like Opprentice face the challenge of imbalanced data, requiring incremental retraining with newly labeled anomalies for better performance.

EGADS~\cite{DBLP:conf/kdd/LaptevAF15}, proposed around the same time as Opprentice~\cite{DBLP:conf/imc/LiuZXSPLJF15}, aimed to enhance classical anomaly detectors.
It consists of three main components: the time-series modeling module (TMM), the anomaly detection module (ADM), and the alerting module (AM).
TMM models the time series and generates expected values for ADM and AM.
EGADS became the first comprehensive system for anomaly detection, as its modules computed prediction errors and filtered out unimportant anomalies.

\subsection{Semi-supervised Methods}
\label{subsec:utad-semi-supervised-methods}

\begin{figure}
    \centering
    \input{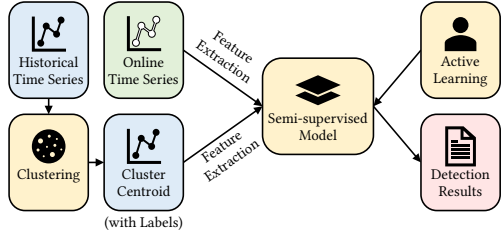}
    \caption{
        The general framework of semi-supervised methods.
    }
    \label{fig:utad-semi-supervised-methods}
\end{figure}

Semi-supervised methods can leverage the precision of supervised learning approaches while simultaneously mitigating the burden of manual labeling by incorporating auxiliary techniques, including clustering, active learning, and transfer learning.
Some common ways of semi-supervised learning are illustrated in Figure~\ref{fig:utad-semi-supervised-methods}.

One of the common approaches is clustering-based semi-supervised learning, which involves clustering UTS based on similarity and then labeling only a representative subset of the clusters.
This approach assumes that anomalous time series will likely be outliers and cannot be grouped into clusters.
By labeling only the representative samples, this method significantly reduces the labeling workload while still maintaining high accuracy.
Label-Less~\cite{DBLP:conf/infocom/ZhaoZLLZP19} stands out as a unique model within this category, designed primarily to expedite the labeling process rather than serving as a dedicated anomaly detection model.
It builds on Opprentice's~\cite{DBLP:conf/imc/LiuZXSPLJF15} feature extraction framework.
The model identifies candidate anomalies and outputs the top-$k$ similar anomalies using Dynamic Time Warping (DTW) as a similarity measure.
This approach eliminates the need for manual search through lengthy time series, significantly reducing operator workload and enhancing efficiency.
Another method, ADS~\cite{DBLP:conf/ipccc/BuLZMLZP18}, utilizes semi-supervised learning by incorporating ROCKA~\cite{DBLP:conf/iwqos/LiZLP18}, an effective clustering method for time series.
Moreover, ADS trains a model using the contrastive pessimistic likelihood estimation (CPLE) algorithm~\cite{DBLP:journals/pami/Loog16}, allowing semi-supervised parameter estimation for any supervised classifier.

Active learning is another approach that reduces the labeling workload by selecting the most informative data samples for labeling.
The model then trains on the labeled data and re-evaluates the unlabeled data to identify the next most informative samples.
This process continues until a satisfactory accuracy level is achieved or the labeling budget is exhausted.
Transfer learning is yet another method that uses pre-trained models to reduce the labeling workload.
In transfer learning, a model trained on a different but related task is adapted to the anomaly detection task by fine-tuning a small labeled dataset.
This approach may result in higher accuracy than training from a blank model due to the transfer of knowledge from the pre-trained model.

ATAD~\cite{DBLP:conf/usenix/ZhangLXQ0QDYCCW19} leverages both active learning and transfer learning techniques.
It extracts general features from labeled and unlabeled time series, including statistical, forecasting, and temporal features.
The method then employs instance-based and feature-based transfer learning methods to train a base detection model.
Active learning recommends informative samples for manual labeling using ``uncertainty'' and ``context diversity'' (to prevent unnecessary labeling), improving the anomaly detector's performance.
ATAD targets scenarios requiring high-precision anomaly detection with limited labeled data, such as cross-dataset tasks.

Although methods like ADS and ATAD have demonstrated promising results, these approaches still require periodic manual label updates.
PUAD~\cite{DBLP:conf/issre/ZhangZSSSZPW21} overcomes these challenges by using Positive-Unlabeled (PU) learning~\cite{DBLP:journals/tmm/ZhangWMTY19}.
It pre-trains a linear model on the positive set (anomalous samples) to produce prediction scores, initializing the negative set.
Next, it employs active-learning-based self-training with a random forest classifier.
The classifier is applied to the unlabeled set, outputting negative and potential positive samples, which are verified by operators.
This process iterates until sufficient labeled samples are obtained.
After obtaining enough labels through PU learning, PUAD trains a semi-supervised model using random-forest-based CPLE~\cite{DBLP:journals/pami/Loog16}, similar to ADS.

\subsection{Unsupervised Methods}
\label{subsec:utad-unsupervised-methods}

This section presents unsupervised methods for UTS anomaly detection in the AIOps domain.
These methods can be broadly categorized into four categories: extreme-value-theory-based methods (\S~\ref{subsubsec:utad-evt-based-methods}), prediction-based methods (\S~\ref{subsubsec:utad-prediction-based-methods}), feature-extraction-based methods (\S~\ref{subsubsec:utad-feature-extraction-based-methods}), and reconstruction-based methods (\S~\ref{subsubsec:utad-reconstruction-based-methods}).

\subsubsection{EVT-Based Methods}
\label{subsubsec:utad-evt-based-methods}

Extreme value theory (EVT)~\cite{beirlant2004statistics} is a statistical branch that deals with extreme deviations from the median of probability distributions.
While various techniques exist to find statistical thresholds, they typically require specific distributional assumptions (\eg Gaussian or exponential) and do not perform well when predicting rare or unprecedented events.
In contrast, EVT infers the distribution of extreme events from a given ordered sample of a random variable and is, therefore, well-suited for such cases.
Recent experiments~\cite{DBLP:conf/kdd/SifferFTL17} have demonstrated the effectiveness of EVT in anomaly detection in the AIOps domain, highlighting its usefulness as a powerful method.

SPOT~\cite{DBLP:conf/kdd/SifferFTL17} is the first application of EVT to UTS anomaly detection, enabling it to avoid assumptions about data distribution or manual threshold tuning.
The peaks-over-threshold (POT) approach fits extreme value distribution and calculates the anomaly threshold.
SPOT handles mid-term seasonality in drifting cases by modeling average local behavior and applying SPOT to relative gaps.
Additionally, SPOT can also serve as an effective tool for automatic thresholding in other anomaly detection methods~\cite{DBLP:journals/jsac/ZhangZLFSZZPSLY22}.

SPOT's shortcomings are twofold: it struggles to detect local anomalies (\S~\ref{subsubsec:point-anomalies}) due to its focus on extreme values in the overall data distribution, and its complex calculations, like the expectation-maximization algorithm, hinder efficiency.
FluxEV~\cite{DBLP:conf/wsdm/LiDSC21} offers a solution with a two-step smoothing process and a method-of-moments (MOM) based threshold calculation.
MOM leads to efficiency improvements of four to six times compared to SPOT.
FluxEV requires only a small amount of training data for startup and initialization, enabling real-time detection with high efficiency.

Another method, StepWise~\cite{DBLP:conf/issre/MaZPHD18}, utilizes EVT in conjunction with an improved singular spectrum transform (iSST)~\cite{DBLP:conf/conext/ZhangLPCQTZ15} to calculate change scores, addressing the concept drift phenomenon in fast software update iterations.
By applying EVT to the change score stream, StepWise automatically sets the threshold.
It employs a robust linear model to fit the relationship between new and old concepts, enabling the seamless operation of old detectors without parameter changes.
It is important to clarify that StepWise is not an anomaly detection model itself; its main purpose is to identify concept drifts and adapt the original anomaly detection model accordingly.

\subsubsection{Prediction-Based Methods}
\label{subsubsec:utad-prediction-based-methods}

UTS anomaly detection methods follow a similar pattern depicted in Figure~\ref{fig:utad-prediction-based-methods}, with the main difference being the prediction model used.
A hybrid forecasting model called ARIMA-WNN is proposed by \citet{DBLP:conf/compsac/QiuDWYC19}.
It combines the linear component of a time series using the ARIMA model and models the nonlinear component with a wavelet neural network (WNN).
Compared to traditional backpropagation networks, WNN has superior fitting abilities and faster convergence for nonlinear functions.
ARIMA-WNN's hybrid approach produces more accurate short-term predictions than other models.

\subsubsection{Feature Extraction-Based Methods}
\label{subsubsec:utad-feature-extraction-based-methods}

Feature-extraction-based methods are also widely used in UTS anomaly detection.
These methods try to extract representative features from time series, capturing their inherent characteristics.
Common features include statistical features (\eg mean, variance), frequency domain features (\eg Fourier transform coefficients), and autocorrelation features (\eg autocorrelation function, partial autocorrelation function).
By distilling time series into feature vectors, the dimensionality of data is reduced while still retaining important properties.
Once features have been extracted, anomaly detection is performed on the feature vectors.
Common techniques for this include clustering-based methods, classification-based methods, and nearest-neighbor-based methods.
Clustering groups normal feature vectors together based on similarity.
Classification trains a model to distinguish normal from abnormal feature vectors.
Nearest neighbor techniques detect anomalies based on the distance between a test feature vector and its closest neighbors in the training set.
These feature extraction-based methods have several benefits.
They are often more computationally efficient since they operate on feature vectors rather than raw time series.
They can also produce better results when the time series is high dimensional.
However, performance relies on extracting features that properly represent normal and abnormal behaviors.

DDCOL~\cite{DBLP:journals/tnsm/YuCWCLL20} is an anomaly detection method that uses feature extraction and density-based clustering techniques.
Unlike most contemporary approaches, DDCOL focuses on the properties of anomalies in addition to the normal pattern of time series.
It extracts and combines high-order difference features to identify anomalies in various time series.
Then, it models the feature distribution of normal data using density-based clustering.
DDCOL achieves real-time and efficient anomaly detection without requiring any anomaly labels, making it suitable for scenarios with many anomalies and high-efficiency demands.

Another feature extraction-based method is SR-CNN~\cite{DBLP:conf/kdd/RenXWYHKXYTZ19}.
It applies a convolutional neural network (CNN) directly to the output of the spectral residual (SR) model.
The SR algorithm involves the Fourier Transform to obtain the log amplitude spectrum, calculation of spectral residual, and inverse Fourier transform to return to the spatial domain.
By using CNN as its discriminative model architecture, SR-CNN simplifies the learning process compared to using the original time series.
It also leverages synthetic anomalies for training, enabling the detector to adapt to changing time-series distributions without manual labeling.
SR-CNN is practical, efficient, and easily integrated into online monitoring systems to provide quick alerts for important metrics.

TS-BERT~\cite{DBLP:conf/iccS/DangZWZYH21} introduces the BERT model, originally designed for natural language processing (NLP), as a feature extractor.
In the pre-training phase, the model learns behavioral features from massive unlabeled data, and then its parameters are fine-tuned with the target dataset.
Three reasons justify using BERT: its core algorithm, Transformer, effectively addresses the long-distance dependence problem; pre-training allows learning from large-scale raw text, reducing reliance on supervised learning and addressing the problem of missing labels; and BERT's comprehensive and open-source nature reduces development costs.

Despite numerous efforts in feature-extraction-based anomaly detection approaches, many existing methods lack interpretability.
These methods calculate probabilities at each timestamp to identify anomalies and then convert them into binary labels using a threshold.
However, this simplistic approach may not satisfy engineers, as they must manually investigate identified metrics for fault localization, which can be challenging for large-scale online services.
In addition, false alerts are common, and many models are trained offline, making them unable to adapt to evolving services and user behaviors due to concept drift.
To address these challenges, ADSketch~\cite{DBLP:conf/icse/ChenL00LL22} is proposed as an interpretable and adaptive approach for service performance anomaly detection.
It characterizes service performance issues using monitoring metrics and provides explanations for its predictions using clustering techniques.
ADSketch can also accept new patterns on the fly, making it adaptable to changing service features and user behaviors.

\subsubsection{Reconstruction-Based Methods}
\label{subsubsec:utad-reconstruction-based-methods}

\begin{figure}
    \centering
    \input{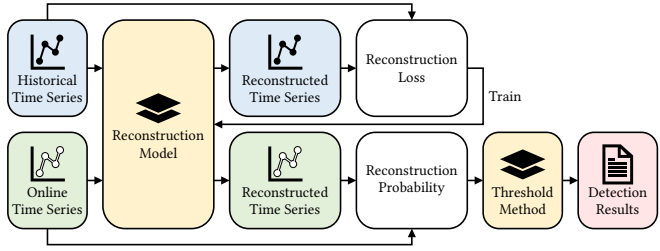}
    \caption{
        The general framework of reconstruction-based methods.
    }
    \label{fig:utad-reconstruction-based-methods}
\end{figure}

Reconstructive methods utilize encoding techniques to compress input data into latent variables, which are subsequently decoded to generate a reconstructed version of the inputs.
This method relies on the assumption that anomalies are more challenging to reconstruct accurately than normal samples.
The general framework of reconstruction-based anomaly detection is depicted in Figure~\ref{fig:utad-reconstruction-based-methods}.
Typically, reconstruction-based methods fall into three categories: contrastive learning, autoencoder (AE), and generative adversarial network (GAN).

Contrastive learning is a successful machine learning technique used in computer vision and NLP.
Recently, it has been applied to anomaly detection, specifically in the TS-CP2 method~\cite{DBLP:conf/www/DeldariSXS21}.
TS-CP2 consists of four parts: negative sampling, representation learning encoder, prediction head, and anomaly detection.
It uses continuous time windows as positive samples and separate time windows as negative samples. The encoder maps time window pairs into compact embeddings, learning similarity on a short time scale through mutual information maximization.
A projection head further reduces dimensionality, and cosine similarity is calculated to identify anomalies.

\begin{figure}
    \centering
    \input{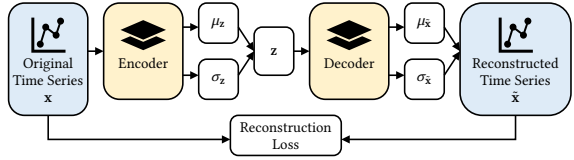}
    \caption{
        The model architecture of VAE.
    }
    \label{fig:utad-vae}
\end{figure}

Another large category of reconstruction-based methods is the autoencoder (AE).
Typically, an AE consists of two key components: an encoder and a decoder.
The encoder encodes input data, while the decoder reconstructs the input by minimizing the difference between the original and output vectors.
High reconstruction error values from a well-trained AE indicate potential anomalies.
However, the AE's latent space may suffer from overfitting and discreteness.
To tackle these issues, the variational autoencoder (VAE) generates two vectors, mean and variance, to sample the latent vector from a distribution.
Figure \ref{fig:utad-vae} illustrates the VAE's model structure.

One of the first VAE-based anomaly detection methods is Donut~\cite{DBLP:conf/www/XuCZLBLLZPFCWQ18}.
Donut employs a three-step process: missing data injection (MDI), training using modified evidence lower bound (ELBO), and detection with Markov chain Monte Carlo (MCMC) based imputation.
MDI is employed to randomly treat normal points as missing, enhancing the robustness of Donut to reconstruct normal points when given anomalous data.
Modified ELBO excludes the contribution of anomalies and missing points during training.
MCMC-based missing data imputation is applied to eliminate biases introduced by missing points.
It's ideal for unsupervised seasonal UTS with local variation, reducing the need for manual tuning and labeling.

Donut, while effective, has a limitation in handling time information-related anomalies commonly found in real-world applications.
To address this, Bagel~\cite{DBLP:conf/ipccc/LiCP18} was introduced as an improved version of Donut.
While sharing architectural similarities with Donut, Bagel utilizes a conditional VAE (CVAE) instead of the standard VAE, allowing it to incorporate timestamp information as a condition variable.
This enables Bagel to handle time information-related anomalies better, including periodic patterns and missing values.
However, both Donut and Bagel assume specific forms of seasonal patterns and zero-mean normally distributed noise in the underlying time series distributions.
Additionally, training both models requires a considerable amount of data and manual threshold setting, posing practical challenges.

ADT-SHL~\cite{DBLP:conf/bdccf/DuanCX19} addresses the efficiency limitations of Donut by combining DTW as the clustering method and Donut's VAE model as the detector.
In addition, ADT-SHL enhances training efficiency by sharing hidden layers.
ADT-SHL employs temporal and spatial convolutional layers in the hidden layers to capture local and global similarities of time series.
When presented with a new time series, ADT-SHL assigns it to the most similar cluster, transfers the anomaly detection model of that cluster, and fine-tunes it to detect anomalies rapidly.
This adaptive approach ensures continued applicability, even with changes in UTS patterns due to software updates.

AnoTransfer~\cite{DBLP:journals/jsac/ZhangZLFSZZPSLY22} adopts a transfer learning approach to enhance anomaly detection efficiency, similar to ADT-SHL.
In the offline training stage, historical UTS undergo clustering and base model training processes.
Each cluster's centroid forms a shape library, and segments closest to the centroid are used to train a CVAE model as the base model for that shape.
In the online environment, when a new or changed UTS arrives, it undergoes the transfer learning stage and fine-tuning with an automatic transfer strategy selection method.
Finally, SPOT~\cite{DBLP:conf/kdd/SifferFTL17} with hyperparameter sharing dynamically calculates the anomaly score threshold.
This transfer-learning approach enables AnoTransfer to efficiently adapt to changes in UTS patterns.

Autoencoders are known for their strong ability to fit data, but they can also overfit anomalous data.
On the other hand, generative adversarial networks (GANs) may struggle to fully capture the hidden distribution of the data, leading to false alarms.
One approach to address this challenge is to improve post-processing steps, which can effectively reduce false positives.
TadGAN~\cite{DBLP:conf/bigdataconf/GeigerLACV20}, based on GAN, computes robust anomaly scores at each time step using GAN's generator and discriminator in a cycle-consistent architecture with time-series-to-time-series mapping.
Furthermore, TadGAN utilizes an anomaly-pruning approach to reduce false positives.

\begin{figure}
    \centering
    \input{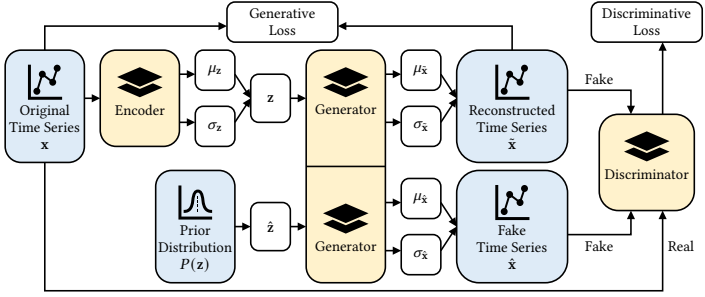}
    \caption{
        The model structure of VAE enhanced GAN.
    }
    \label{fig:utad-vae-gan}
\end{figure}

GAN-based methods encounter the challenge of solving the inverse mapping problem, where finding the best mapping from the original to the latent space is difficult due to the non-invertibility of the mapping from the latent to the original space.
That is, multiple latent vectors can generate the same time series in the original space.
An anomaly detection method called VAE-GAN was proposed in~\cite{DBLP:journals/sensors/NiuYW20} to address this issue.
VAE-GAN combines VAE as the encoder and GAN as the decoder to map data between the latent and original spaces.
Jointly training the encoder, generator, and discriminator enhances VAE-GAN with better anomaly distinction by combining reconstruction differences and discriminator results.
The model structure of this idea is illustrated in Figure~\ref{fig:utad-vae-gan}.
Other methods like Buzz~\cite{DBLP:conf/infocom/ChenXLPCQFW19}, Adran~\cite{DBLP:journals/tnsm/NieZL21} and IoT-GAN~\cite{DBLP:conf/ica3pp/ChenZJCHG21} also take similar approaches.

\subsection{Summary}
\label{subsec:utad-summary}

In summary, early classical methods for UTS anomaly detection in the AIOps domain were mainly applied to network traffic and relied on pre-defined rules or simple statistical detectors.
Some of these methods started incorporating machine learning techniques by combining multiple detector outputs.
mong supervised methods, Opprentice and EGADS stood out, which were feature-extraction-based and prediction-based anomaly detection frameworks, respectively.
To reduce labeling workload while maintaining model performance, semi-supervised methods introduced techniques like clustering, active learning, and transfer learning.
However, the increasing data volume made acquiring sufficient labels challenging, leading to a growing need for unsupervised learning.
Unsupervised methods often rely on deep generative models like VAE and GAN, while some combine deep neural networks with traditional techniques like EVT and SR.

Table~\ref{tab:utad-summary} presents a comparison of various UTS anomaly detection methods across different perspectives.
The methods are categorized into three groups: prediction-based (P), reconstruction-based (R), and ensemble (E) methods.
Furthermore, they are classified as unsupervised (U), supervised (S), or semi-supervised (Semi) based on their need for labeled data during training.
In the \textit{Noise Assumption} column, \ding{51} indicates that the method assumes Gaussian-distributed noise in the data, while \ding{55} signifies that the method makes no assumptions about the noise distribution.
The table includes each method's primary technique, public evaluation datasets, and evaluation metrics. Commonly used evaluation datasets and their links are listed in Table~\ref{tab:utad-datasets}.
For additional information on evaluation metrics, please refer to \S~\ref{sec:evaluation-metrics}.

\begin{table}
    \centering
    \tiny
    \begin{threeparttable}
        \caption{
            Summary of UTS Anomaly Detection Methods in the AIOps Domain
        }
        \label{tab:utad-summary}
        \begin{tabular}{lccccc}
            \toprule
            Method                                                  & Type   & Techniques          & Noise Assumption & Evaluation Datasets & Evaluation Metrics\tnote{*} \\
            \midrule
            \citet{DBLP:conf/infocom/LeePHPLYGYK12}                 & P/U    & Historical Average  & \ding{51}        & \textemdash         & FPR/FNR                     \\
            \citet{DBLP:conf/sigcomm/ChoffnesBG10}                  & P/U    & MA                  & \ding{55}        & \textemdash         & FPR                         \\
            \citet{DBLP:conf/imc/KrishnamurthySZC03}                & P/U    & EWMA                & \ding{55}        & \textemdash         & FPR/FNR                     \\
            \citet{DBLP:conf/imc/ZhangGGR05}                        & P/U    & ARIMA               & \ding{55}        & \textemdash         & \textemdash                 \\
            \citet{DBLP:conf/icc/HimuraFCE09}                       & P/U    & MSGF                & \ding{55}        & (h)                 & Accuracy                    \\
            Argus~\cite{DBLP:conf/infocom/YanFGGMPSY12}             & P/U    & Holt-Winters        & \ding{55}        & \textemdash         & \textemdash                 \\
            PRISM~\cite{DBLP:conf/conext/MahimkarGWYZEHS11}         & P/U    & SVD                 & \ding{55}        & \textemdash         & ROC                         \\
            \citet{DBLP:conf/imc/BarfordKPR02}                      & P/U    & Wavelet Filter      & \ding{55}        & \textemdash         & Recall                      \\
            MAWILab~\cite{DBLP:conf/conext/FontugneBAF10}           & E/U    & Majority Vote       & \ding{55}        & (h)                 & \textemdash                 \\
            PAD~\cite{DBLP:journals/network/ShanbhagW09}            & E/U    & Normalization       & \ding{55}        & \textemdash         & FPR/FNR                     \\
            SDnEA~\cite{DBLP:conf/icc/AshfaqJKR10}                  & E/S    & Entropy             & \ding{55}        & \textemdash         & Recall                      \\
            aHSIC~\cite{DBLP:conf/ijcai/YamadaKNS13}                & P/S    & HSIC                & \ding{55}        & \textemdash         & AUC                         \\
            \citet{DBLP:conf/sigcomm/ChenMSZ13}                     & P/U    & TSD                 & \ding{51}        & \textemdash         & FPR/FNR                     \\
            MERCURY~\cite{DBLP:conf/sigcomm/MahimkarSGSWYZE10}      & P/U    & CUSUM               & \ding{55}        & \textemdash         & FPR/FNR                     \\
            \citet{DBLP:conf/imc/SouleST05}                         & P/U    & CUSUM/GLR           & \ding{51}        & \textemdash         & ROC                         \\
            \midrule
            Opprentice~\cite{DBLP:conf/imc/LiuZXSPLJF15}            & E/S    & Random Forest       & \ding{55}        & \textemdash         & $F_{1}$/AUC                 \\
            EGADS~\cite{DBLP:conf/kdd/LaptevAF15}                   & P/S    & \textemdash         & \ding{51}        & (a)                 & $F_{1}$                     \\
            \midrule
            Label-less~\cite{DBLP:conf/infocom/ZhaoZLLZP19}         & E/Semi & Isolation Forest    & \ding{55}        & \textemdash         & $F_{1}$/AUC                 \\
            ADS~\cite{DBLP:conf/ipccc/BuLZMLZP18}                   & E/Semi & Random Forest       & \ding{55}        & \textemdash         & Point-Adjusted $F_{1}$      \\
            ATAD~\cite{DBLP:conf/usenix/ZhangLXQ0QDYCCW19}          & E/Semi & Random Forest       & \ding{55}        & (a)/(b)             & $F_{1}$                     \\
            PUAD~\cite{DBLP:conf/issre/ZhangZSSSZPW21}              & E/Semi & Random Forest       & \ding{55}        & (b)                 & Point-Adjusted $F_{1}$      \\
            \midrule
            SPOT~\cite{DBLP:conf/kdd/SifferFTL17}                   & P/U    & EVT                 & \ding{55}        & (h)                 & ROC                         \\
            FluxEV~\cite{DBLP:conf/wsdm/LiDSC21}                    & P/U    & EVT                 & \ding{55}        & (a)/(c)             & Point-Adjusted $F_{1}$      \\
            StepWise~\cite{DBLP:conf/issre/MaZPHD18}                & P/U    & Robust Linear Model & \ding{55}        & \textemdash         & $F_{1}$                     \\
            ARIMA-WNN~\cite{DBLP:conf/compsac/QiuDWYC19}            & P/U    & WNN                 & \ding{51}        & \textemdash         & MAE/RMSE/MAPE               \\
            DDCOL~\cite{DBLP:journals/tnsm/YuCWCLL20}               & E/U    & DBSCAN              & \ding{51}        & (a)/(c)             & Point-Adjusted $F_{1}$      \\
            SR-CNN~\cite{DBLP:conf/kdd/RenXWYHKXYTZ19}              & E/U    & SR/CNN              & \ding{51}        & (a)/(c)             & Point-Adjusted $F_{1}$      \\
            TS-BERT~\cite{DBLP:conf/iccS/DangZWZYH21}               & E/U    & SR/BERT             & \ding{51}        & (a)/(c)             & Point-Adjusted $F_{1}$      \\
            ADSketch~\cite{DBLP:conf/icse/ChenL00LL22}              & E/U    & STAMP/AP            & \ding{51}        & (a)/(c)             & $F_{1}$                     \\
            TS-CP2~\cite{DBLP:conf/www/DeldariSXS21}                & R/U    & CNN/MLP             & \ding{55}        & (a)/(f)/(g)         & Point-Adjusted $F_{1}$      \\
            Donut~\cite{DBLP:conf/www/XuCZLBLLZPFCWQ18}             & R/U    & VAE                 & \ding{51}        & \textemdash         & Point-Adjusted $F_{1}$      \\
            Bagel~\cite{DBLP:conf/ipccc/LiCP18}                     & R/U    & CVAE                & \ding{51}        & \textemdash         & Point-Adjusted $F_{1}$      \\
            AnoTransfer~\cite{DBLP:journals/jsac/ZhangZLFSZZPSLY22} & R/U    & CVAE                & \ding{51}        & (b)                 & Point-Adjusted $F_{1}$      \\
            ADT-SHL~\cite{DBLP:conf/bdccf/DuanCX19}                 & R/U    & VAE                 & \ding{51}        & (a)/(d)/(e)         & $F_{1}$                     \\
            Buzz~\cite{DBLP:conf/infocom/ChenXLPCQFW19}             & R/U    & GAN/VAE             & \ding{51}        & \textemdash         & Point-Adjusted $F_{1}$/AUC  \\
            TadGAN~\cite{DBLP:conf/bigdataconf/GeigerLACV20}        & R/U    & GAN/AE              & \ding{51}        & (a)/(b)             & Range-Based $F_{1}$         \\
            LSTM-based VAE-GAN~\cite{DBLP:journals/sensors/NiuYW20} & R/U    & RNN/GAN/VAE         & \ding{51}        & (a)/(c)             & $F_{1}$                     \\
            Adran~\cite{DBLP:journals/tnsm/NieZL21}                 & R/U    & RAN                 & \ding{55}        & (c)                 & $F_{1}$                     \\
            IoT-GAN~\cite{DBLP:conf/ica3pp/ChenZJCHG21}             & R/U    & GAN/VE              & \ding{51}        & (a)                 & Range-Based $F_{1}$         \\
            \bottomrule
        \end{tabular}
        \begin{tablenotes}
            \item [*] FPR: False Positive Rate; FNR: False Negative Rate; AUC: Area Under the Curve; ROC: Receiver Operating Characteristic; MAE: Mean Absolute Error; RMSE: Root Mean Square Error; MAPE: Mean Absolute Percentage Error.
        \end{tablenotes}
    \end{threeparttable}
\end{table}

\begin{table}
    \centering
    \tiny
    \caption{
        Public UTS Anomaly Detection Datasets in the AIOps Domain Sorted by Frequency of Usage
    }
    \label{tab:utad-datasets}
    \begin{tabular}{ccl}
        \toprule
        No. & Name         & Link                                                                         \\
        \midrule
        (a) & Yahoo        & \url{https://webscope.sandbox.yahoo.com/}                                    \\
        (b) & NAB          & \url{https://github.com/numenta/NAB}                                         \\
        (c) & AIOPS (2018) & \url{http://iops.ai/dataset_list/}                                           \\
        (d) & Baidu        & \url{https://github.com/baidu/Curve}                                         \\
        (e) & Alibaba      & \url{https://github.com/alibaba/clusterdata/tree/master/cluster-trace-v2018} \\
        (f) & USC-HAD      & \url{http://sipi.usc.edu/had}                                                \\
        (g) & HASC         & \url{http://hasc.jp/hc2011}                                                  \\
        (h) & MAWI         & \url{https://mawi.wide.ad.jp/mawi/}                                          \\
        \bottomrule
    \end{tabular}
\end{table}

Table~\ref{tab:utad-repos} provides a summary of the source codes for UTS anomaly detection methods, facilitating easy reproduction and deployment.
It includes official code repositories for most methods, but some may either be unofficial or not publicly accessible.
Methods without publicly available codes are excluded from the table.

\begin{table}
    \centering
    \tiny
    \caption{
        Source Code of UTS Anomaly Detection Datasets in the AIOps Domain
    }
    \label{tab:utad-repos}
    \begin{tabular}{cccc}
        \toprule
        Name        & Language & Method                                       & Code                                                      \\
        \midrule
        EGADS       & Java     & \citet{DBLP:conf/kdd/LaptevAF15}             & \url{https://github.com/yahoo/egads}                      \\
        PUAD        & Python   & \citet{DBLP:conf/issre/ZhangZSSSZPW21}       & \url{https://github.com/PUAD-code/PUAD}                   \\
        SPOT        & Python   & \citet{DBLP:conf/kdd/SifferFTL17}            & \url{https://github.com/Amossys-team/SPOT}                \\
        SR-CNN      & Python   & \citet{DBLP:conf/kdd/RenXWYHKXYTZ19}         & \url{https://github.com/y-bar/ml-based-anomaly-detection} \\
        ADSketch    & Python   & \citet{DBLP:conf/icse/ChenL00LL22}           & \url{https://github.com/opspai/adsketch}                  \\
        TS-CP2      & Python   & \citet{DBLP:conf/www/DeldariSXS21}           & \url{https://github.com/cruiseresearchgroup/TSCP2}        \\
        Donut       & Python   & \citet{DBLP:conf/www/XuCZLBLLZPFCWQ18}       & \url{https://github.com/NetManAIOps/donut}                \\
        Bagel       & Python   & \citet{DBLP:conf/ipccc/LiCP18}               & \url{https://github.com/NetManAIOps/Bagel}                \\
        AnoTransfer & Python   & \citet{DBLP:journals/jsac/ZhangZLFSZZPSLY22} & \url{https://github.com/anotransfer/AnoTransfer-code}     \\
        Buzz        & Python   & \citet{DBLP:conf/infocom/ChenXLPCQFW19}      & \url{https://github.com/yantijin/Buzz}                    \\
        TadGAN      & Python   & \citet{DBLP:conf/bigdataconf/GeigerLACV20}   & \url{https://github.com/sintel-dev/Orion}                 \\
        \bottomrule
    \end{tabular}
\end{table}

\section{Multivariate Time Series}
\label{sec:multivariate-time-series}
In addition to analyzing univariate time series (UTS), the monitoring of a complex system often requires the collective assessment of multiple metrics to describe its overall state.
Consequently, detecting anomalies in multivariate time series (MTS) becomes necessary to trigger system-level alarms.
This section focuses on various methods to detect anomalies in MTS within the AIOps domain.

The MTS anomaly detection problem differs from its univariate counterpart due to correlations among features/metrics.
It uses an input matrix $\mathbf{X}$ with dimensions $M\times W$, where $M$ is the number of features and $W$ is the window size (see Section~\ref{subsec:input-data}).
The output is an anomaly score, quantifying the likelihood of a data sequence or data point being anomalous.
MTS methods incorporate models for spatial and temporal dependence and can collaborate in detection.
In certain scenarios, the features extracted by spatial dependence models can serve as input for temporal dependence models.
While some explicitly model spatiotemporal correlations, others treat each feature independently and aggregate anomaly scores based on temporal dependence.
These concepts will be further discussed.

Furthermore, rather than using reconstruction-based and prediction-based models separately, it's possible to integrate both types within a single MTS anomaly detection method.
This review aims to categorize models based on their capacity to recover the original time series.
Additionally, the study compiles several publicly available datasets commonly used in MTS anomaly detection (see Table~\ref{tab:mtad-datasets}) sorted by their frequency of usage.
The table also includes information on how to access these datasets.

\begin{figure}
    \centering
    \input{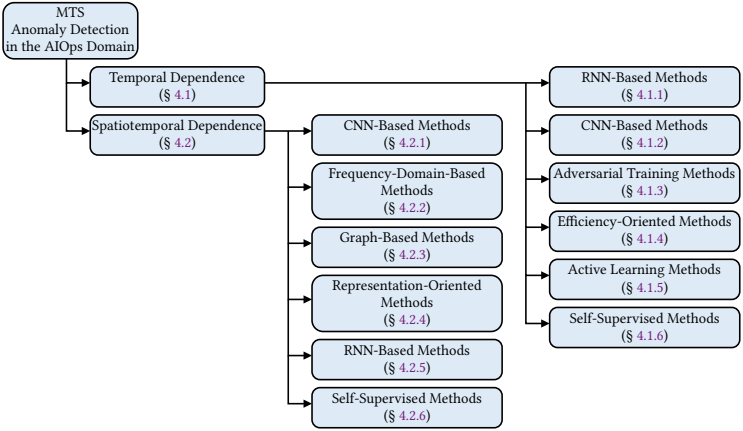}
    \caption{
        Types of methods for detecting anomalies in MTS.
    }
    \label{fig:mtad-overview}
\end{figure}

The organization of this section is depicted in Figure~\ref{fig:mtad-overview}.
In terms of model implementation, MTS anomaly detection approaches can be broadly classified into two categories: temporal dependence learning (\S~\ref{subsec:mtad-temporal-dependence}) and spatiotemporal dependence learning (\S~\ref{subsec:mtad-spatiotemporal-dependence}) methods.
Furthermore, within these two categories, there are further specific classifications based on the implementation of their underlying frameworks.

\subsection{Temporal Dependence}
\label{subsec:mtad-temporal-dependence}

In the early years, many MTS anomaly detection methods concentrated on capturing temporal features.
Most of these approaches utilized recurrent neural networks (RNNs) (\S~\ref{subsubsec:mtad-temporal-dependence-rnn-based-methods}) or convolutional neural networks (CNNs) (\S~\ref{subsubsec:mtad-temporal-dependence-cnn-based-methods})to effectively model the complex temporal dependence within the MTS data.
Some methods incorporated advanced training techniques, such as adversarial training (\S~\ref{subsubsec:mtad-temporal-dependence-adversarial-training-methods}), active learning (\S~\ref{subsubsec:mtad-temporal-dependence-active-learning-methods}), and self-supervised learning (\S~\ref{subsubsec:mtad-temporal-dependence-self-supervised-methods}) to enhance the robustness of the model.
The efficiency of these methods was also improved by employing various techniques (\S~\ref{subsubsec:mtad-temporal-dependence-efficiency-oriented-methods}).

\subsubsection{RNN-Based Methods}
\label{subsubsec:mtad-temporal-dependence-rnn-based-methods}

One of the most straightforward approaches involves the integration of autoencoders (AEs) with RNNs.
RNNs are employed as independent encoders and decoders, facilitating sequence-to-sequence projection.
Typically, the RNN receives standardized MTS or embedding vectors from preceding components.
Through the temporal dimension projection, the RNN's output effectively preserves the global information of the original time series window.

In the AIOps domain, like network traffic monitoring systems, the dynamic emergence of new workloads requires anomaly detection models to recognize previously unseen patterns, possibly forgetting previous ones, and respond adequately to feedback from system administrators.
To address these challenges, UNLEARN~\cite{DBLP:conf/ccs/DuCLOS19} combines Long Short-Term Memory (LSTM) and AE to capture temporal dependence within MTS.
UNLEARN focuses on mitigating exploding loss and catastrophic forgetting using a special objective function:
\begin{equation}
    \mathcal{L}_{\mathrm{UNLEARN}}\left(\mathbf{X}\right)=-\sum_{t}l_{t}\mathcal{L}\left(\mathbf{x}_{t}\right)
\end{equation}
Here, $\mathcal{L}$ measures the reconstruction error, and $l_{t}$ indicates the prediction result.
Specifically, $l_{t}=1$ if the prediction of $\mathbf{x}_{t}$ is a true positive, and $-1$ if the prediction of $\mathbf{x}_{t}$ is a false negative.
The objective function can be revised with a lower bound to address exploding loss.
UNLEARN also incorporates feedback from system operators, indicating whether $\mathbf{x}_{t}$ has been falsely predicted, allowing timely adaptive adjustments.

To overcome the limitations of deterministic methods like RNNs in learning robust representations of MTS, certain approaches integrate RNNs and VAEs to account for both temporal dependence and the stochastic nature of MTS.
This architecture is known as a recurrent VAE, as demonstrated by OmniAnomaly~\cite{DBLP:conf/kdd/SuZNLSP19}.
Another effort to address the challenge is the time-series memory autoencoder (TSMAE) proposed by \citet{gao2022tsmae}.
TSMAE combines AE with a memory module based on the LSTM model, preserving expressive capabilities without overfitting. The model's loss function considers both the reconstruction loss and the memory sparsity penalty, thus promoting the extraction of patterns using fewer memory terms.

To address the issue of computational complexity, LR-SemiVAE~\cite{DBLP:journals/apin/ChenTDHG23} introduces the parallel multi-head dynamic attention (MDA) mechanism within the LSTM framework.
This approach accelerates the training process by enabling parallel processing of individual time series.

\subsubsection{CNN-Based Methods}
\label{subsubsec:mtad-temporal-dependence-cnn-based-methods}

RNNs are known for their high computational demands and lengthy training time.
In contrast, CNNs have distinct features such as parameter sharing and permutation invariance, reducing complexity and facilitating robust representation learning.
CAE-Ensemble~\cite{DBLP:journals/pvldb/CamposKGHZYJ21} improves computational efficiency by using a convolutional AE as its foundation and an ensemble of multiple basic models.
Additionally, a diversity metric assesses the similarity between each pair of basic models, $f_{m}$ and $f_{n}$, using the equation:
\begin{equation}
    DIV_{f_{m},f_{n}}\left(\mathbf{X}\right)=\lVert f_{m}\left(\mathbf{X}\right)-f_{n}\left(\mathbf{X}\right)\rVert_{2}
\end{equation}

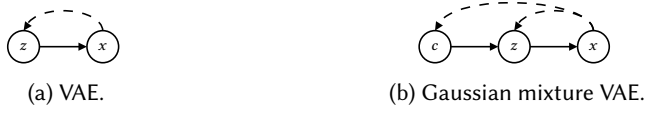
\begin{figure}
    \centering
    \begin{subfigure}{0.4\linewidth}
        \centering
        \tikzset{every picture/.style={line width=0.75pt}} 

\scalebox{\tikzscale}{\begin{tikzpicture}[x=0.75pt,y=0.75pt,yscale=-1,xscale=1]

\draw    (20,22.43) -- (47,22.43) ;
\draw [shift={(50,22.43)}, rotate = 180] [fill={rgb, 255:red, 0; green, 0; blue, 0 }  ][line width=0.08]  [draw opacity=0] (5.36,-2.57) -- (0,0) -- (5.36,2.57) -- cycle    ;
\draw   (0,22.43) .. controls (0,16.91) and (4.48,12.43) .. (10,12.43) .. controls (15.52,12.43) and (20,16.91) .. (20,22.43) .. controls (20,27.96) and (15.52,32.43) .. (10,32.43) .. controls (4.48,32.43) and (0,27.96) .. (0,22.43) -- cycle ;
\draw   (50,22.43) .. controls (50,16.91) and (54.48,12.43) .. (60,12.43) .. controls (65.52,12.43) and (70,16.91) .. (70,22.43) .. controls (70,27.96) and (65.52,32.43) .. (60,32.43) .. controls (54.48,32.43) and (50,27.96) .. (50,22.43) -- cycle ;
\draw  [dash pattern={on 4.5pt off 4.5pt}]  (60,12.43) .. controls (59.05,-1.56) and (20.01,-5.51) .. (11.14,9.85) ;
\draw [shift={(10,12.43)}, rotate = 304.81] [fill={rgb, 255:red, 0; green, 0; blue, 0 }  ][line width=0.08]  [draw opacity=0] (5.36,-2.57) -- (0,0) -- (5.36,2.57) -- cycle    ;

\draw (10,22.43) node  [font=\scriptsize] [align=left] {$\displaystyle z$};
\draw (60,22.43) node  [font=\scriptsize] [align=left] {$\displaystyle x$};

\end{tikzpicture}}
        \caption{
            VAE.
        }
    \end{subfigure}
    \quad
    \begin{subfigure}{0.4\linewidth}
        \centering
        \tikzset{every picture/.style={line width=0.75pt}} 

\scalebox{\tikzscale}{\begin{tikzpicture}[x=0.75pt,y=0.75pt,yscale=-1,xscale=1]

\draw    (70,27.43) -- (97,27.43) ;
\draw [shift={(100,27.43)}, rotate = 180] [fill={rgb, 255:red, 0; green, 0; blue, 0 }  ][line width=0.08]  [draw opacity=0] (5.36,-2.57) -- (0,0) -- (5.36,2.57) -- cycle    ;
\draw   (50,27.43) .. controls (50,21.91) and (54.48,17.43) .. (60,17.43) .. controls (65.52,17.43) and (70,21.91) .. (70,27.43) .. controls (70,32.96) and (65.52,37.43) .. (60,37.43) .. controls (54.48,37.43) and (50,32.96) .. (50,27.43) -- cycle ;
\draw   (100,27.43) .. controls (100,21.91) and (104.48,17.43) .. (110,17.43) .. controls (115.52,17.43) and (120,21.91) .. (120,27.43) .. controls (120,32.96) and (115.52,37.43) .. (110,37.43) .. controls (104.48,37.43) and (100,32.96) .. (100,27.43) -- cycle ;
\draw  [dash pattern={on 4.5pt off 4.5pt}]  (110,17.43) .. controls (109.05,3.44) and (70.01,-0.51) .. (61.14,14.85) ;
\draw [shift={(60,17.43)}, rotate = 304.81] [fill={rgb, 255:red, 0; green, 0; blue, 0 }  ][line width=0.08]  [draw opacity=0] (5.36,-2.57) -- (0,0) -- (5.36,2.57) -- cycle    ;
\draw    (20,27.43) -- (47,27.43) ;
\draw [shift={(50,27.43)}, rotate = 180] [fill={rgb, 255:red, 0; green, 0; blue, 0 }  ][line width=0.08]  [draw opacity=0] (5.36,-2.57) -- (0,0) -- (5.36,2.57) -- cycle    ;
\draw   (0,27.43) .. controls (0,21.91) and (4.48,17.43) .. (10,17.43) .. controls (15.52,17.43) and (20,21.91) .. (20,27.43) .. controls (20,32.96) and (15.52,37.43) .. (10,37.43) .. controls (4.48,37.43) and (0,32.96) .. (0,27.43) -- cycle ;
\draw  [dash pattern={on 4.5pt off 4.5pt}]  (110,17.43) .. controls (109.04,-2.5) and (22.21,-7.89) .. (10.92,14.85) ;
\draw [shift={(10,17.43)}, rotate = 301.87] [fill={rgb, 255:red, 0; green, 0; blue, 0 }  ][line width=0.08]  [draw opacity=0] (5.36,-2.57) -- (0,0) -- (5.36,2.57) -- cycle    ;

\draw (60,27.43) node  [font=\scriptsize] [align=left] {$\displaystyle z$};
\draw (110,27.43) node  [font=\scriptsize] [align=left] {$\displaystyle x$};
\draw (10,27.43) node  [font=\scriptsize] [align=left] {$\displaystyle c$};

\end{tikzpicture}}
        \caption{
            Gaussian mixture VAE.
        }
    \end{subfigure}
    \caption{
        The traditional VAE and the Gaussian mixture VAE.
        $x$ is the input, $z$ is the stochastis latent variable, and $c$ is the categorical latent variable.
        The solid lines depict the generative process, while the dashed lines represent the variational approximation.
    }
    \label{fig:mtad-gmvae}
\end{figure}

Another method, DOMI~\cite{DBLP:journals/tc/SuZSZWZLLTWP22}, stands as the pioneering unsupervised model within the AIOps domain that combines CNN-VAE architecture with a Gaussian mixture component.
This integration enables DOMI to effectively leverage complex temporal features and address the challenge of shape diversity.
Figure~\ref{fig:mtad-gmvae} illustrates the distinction between the traditional VAE and the Gaussian mixture VAE.

\subsubsection{Adversarial Training Methods}
\label{subsubsec:mtad-temporal-dependence-adversarial-training-methods}

As discussed earlier, AEs and VAEs are commonly used for anomaly detection by evaluating dissimilarity between input data and reconstructed outputs.
However, their effectiveness diminishes with anomalies in the input.
To handle such situations, a method called USAD~\cite{DBLP:conf/kdd/AudibertMGMZ20} leverages adversarial training in an encoder-decoder architecture.
The general adversarial training architecture for MTS anomaly detection is very similar to that of the univariate counterpart (Figure~\ref{fig:utad-vae-gan}).
However, USAD includes an additional discriminator in its architecture.
This discriminator helps prioritize the reconstruction error of anomalous inputs.
Moreover, USAD introduces a novel two-phase reconstruction mechanism, implementing both the discriminator and generator with AEs, which improves generalization for various normal patterns.

MAD-GAN~\cite{DBLP:conf/icann/LiCJSGN19} presents another GAN-based method.
It combines GAN with LSTM to capture temporal dependence.
MAD-GAN introduces a novel anomaly score called DR-Score (DRS) to detect anomalies.
The DRS incorporates reconstruction loss and discrimination loss and maps the anomaly detection loss of subsequences back to the original time series.
It is defined as:
\begin{equation}
    \begin{aligned}
        DRS_{t} & =\frac{\sum_{j,s\in{j+s=t}}L_{j,s}}{lc_{t}} \\
        lc_{t}  & =count\left(j,s\in{j+s=t}\right)
    \end{aligned}
\end{equation}
Here, $t$ is the index of the original time series, and $L_{j,s}$ represents the anomaly detection loss for a specific subsequence.

To address the challenge of handling unseen patterns and the absence of labels, \citet{DBLP:conf/icde/ChenDHZZZZ21} propose a novel unsupervised anomaly detection framework called DAEMON.
This framework utilizes adversarial training and optimization with three objectives, employing two discriminators.
Similar to the two-phase generation approach in USAD, DAEMON is trained using a two-player min-max game.
In the offline training stage, DAEMON uses 1D CNN to encode the MTS.
Instead of directly decoding the latent variable, it assumes a prior distribution for the latent vector.
Subsequently, adversarial training is applied to enhance generalization even for unseen patterns.

\subsubsection{Efficiency-Oriented Methods}
\label{subsubsec:mtad-temporal-dependence-efficiency-oriented-methods}

In certain application scenarios, computational efficiency has become a significant concern.
The dimension explosion problem in many MTS anomaly detection methods leads to extended initialization times.
To address this challenge, efficiency-oriented models often employ transfer learning or shape-based clustering techniques to accomplish their goals.

JumpStarter~\cite{DBLP:conf/usenix/MaZ0XLLNZWP21} uses hierarchical clustering and compressed sensing (CS) to quickly initialize MTS anomaly detection.
It employs shape-based clustering to group UTS within MTS into clusters during offline processing.
An innovative outlier-resistant sampling algorithm is applied to each group of UTS to address sampling from anomalous segments.
Then, the CS technique reconstructs the groups of sampled UTS.

CTF~\cite{DBLP:conf/infocom/SunSZCLPWZLT21} proposes a model transfer-based framework that employs shape-based hierarchical clustering to address the dimension explosion problem.
Specifically, CTF utilizes a pre-trained coarse-grained model to extract and compress per-machine features into a distribution.
Machines are clustered based on this distribution, and fine-tuning is performed using per-cluster models.
CTF enhances overall training efficiency by leveraging clustering on the per-machine latent representation distribution and reusing partial-layer model fine-tuning.

\citet{DBLP:journals/asc/LiIPJ21} also incorporate shape-based clustering in their approach.
They employ a fixed-length sliding window to divide long MTS into shorter subsequences and extend the Fuzzy C-Means clustering (FCM) algorithm to learn the underlying structure.
By applying a new reconstruction criterion and particle swarm optimization, the anomaly score of each subsequence is computed based on the obtained clusters.
This anomaly detection framework is designed to identify anomalous amplitude and shape patterns in the AIOps domain.

The key idea behind these methods is the utilization of a weight-sharing policy for training MTS with similar patterns.
These approaches commonly use transfer learning in two-phase anomaly detection scenarios: offline training and online detection.
However, methods with shape-based clustering components are more sensitive to anomalies, as they assume that the patterns observed in the training data are normal.
They may have a narrower application scope, requiring a trade-off between computational efficiency and generalization abilities.

There are also methods that specifically address the dynamic and evolving nature of MTS.
These methods are tailored to effectively handle concept drift, a prevalent phenomenon in MTS within the AIOps domain.
ADTCD~\cite{xu2023adtcd} is an advanced anomaly detection approach for streaming Internet of Things (IoT) data.
It uses knowledge distillation, where a teacher model imparts knowledge to a continuously updated student model.
This process keeps the student model up-to-date with changing data.
ADTCD also uses a one-class SVM outlier removal method to prevent model self-poisoning.
The combination of knowledge distillation and dynamic weight adjustment ensures timely and accurate anomaly detection under concept drift.

\subsubsection{Active Learning Methods}
\label{subsubsec:mtad-temporal-dependence-active-learning-methods}

Traditional MTS anomaly detection methods often require a large number of labeled samples to achieve desirable performance.
Consequently, some methods employ active learning to efficiently train their models with limited labels.
By using an active learning technique, an anomaly detection method can determine which samples are most valuable for expert annotation, thus reducing the number of samples that need to be labeled.

Active-MTSAD~\cite{DBLP:conf/dsn/WangCXH22} presents an anomaly detection framework based on active learning.
The unsupervised anomaly detection module generates anomaly scores, used to select important samples for expert labeling, facilitating efficient expert-model interaction.
Active-MTSAD utilizes denominator penalty, negative penalty, and metric learning in the active learning module to detect new anomalous patterns with unseen data distributions.
For query strategy, ensemble methods are employed to compute a candidate set of samples, chosen for expert labeling based on perceived value.

\subsubsection{Self-Supervised Methods}
\label{subsubsec:mtad-temporal-dependence-self-supervised-methods}

Unsupervised methods do not require user feedback or labels.
This lack of supervision can sometimes lead to poor performance in practical applications.
In order to address this limitation, some approaches have turned to the successful applications of contrastive learning, which enables the learning of transformation-invariant representations.
Two such methods are CAE-AD~\cite{DBLP:journals/isci/ZhouYZWY22} and ACVAE~\cite{DBLP:journals/jsac/LiZGZWCJVSP22}.
We will provide details about CAE-AD in this subsection, while ACVAE will be described in \S~\ref{subsec:mtad-spatiotemporal-dependence}.

CAE-AD is a method that combines AE with contrastive learning to create a robust representation of MTS.
It introduces innovative data augmentation techniques in the time and frequency domains to capture invariant characteristics of normal patterns that evolve over time.
Additionally, CAE-AD uses multi-grained contrasting to facilitate robust representation learning.
Contextual contrasting is employed to retain position information that the self-attention mechanism may neglect.
Instance contrasting is utilized to capture the local invariant characteristics of MTS.

\subsection{Spatiotemporal Dependence}
\label{subsec:mtad-spatiotemporal-dependence}

The interplay between multiple metrics in a complex system, characterized by consistent behaviors, is crucial for learning robust representations of MTS.
This correlation among metrics, referred to as spatial dependence, holds significant importance.
This section introduces several approaches that capture features across both temporal and spatial dimensions.
By considering the spatial dependence, they can effectively exploit the collective behavior of multiple metrics to improve the overall anomaly detection performance.

\subsubsection{CNN-Based Methods}
\label{subsubsec:mtad-spatiotemporal-dependence-cnn-based-methods}

MSCRED~\cite{DBLP:conf/aaai/ZhangSCFLCNZCC19}, short for multi-scale convolutional recurrent encoder-decoder, employs a convolutional encoder to capture inter-metric correlation patterns in multi-scale signature matrices.
The signature matrix is obtained by calculating the pairwise inner product of two UTS.
It uses attention-based ConvLSTM to reconstruct the signature matrices, allowing end-to-end processing.
As a result, MSCRED simultaneously addresses anomaly detection, root cause identification, and anomaly severity interpretation.
The signature matrix technique can also significantly aid CNNs in extracting correlations in MTS.
MTS-DCGAN~\cite{DBLP:journals/ijon/LiangSWGLL21}, a multi-time scale deep convolutional generative network, improves upon MSCRED in extracting correlations in MTS.
It generates multi-channel signature matrices using multi-time scale sliding windows and employs a forgetting mechanism to reduce the impact of older data.
Additionally, MTS-DCGAN uses adversarial training with convolutional structures to capture non-linear cross-correlation features.
It introduces a novel threshold-setting strategy for effective anomaly prediction and addresses imbalanced normal and abnormal data.
Both MSCRED and MTS-DCGAN aim to convert the original MTS into 2D matrices, enabling similar processing to 2D images.

SLA-VAE~\cite{DBLP:conf/www/HuangCL22} employs convolutions in the spatial dimension to capture inter-metric dependence and utilizes active learning to update the VAE model using a small number of uncertain samples.
SLA-VAE enables model sharing among entities of the same type, thereby reducing training time.
Similarly, PBATPC~\cite{DBLP:journals/iotj/HanLPLC22}, which stands for multivariate-time-series-prediction adaptive data transmission period control algorithm, utilizes a 2D convolution kernel to simultaneously handle timing and spatial dependence.
It employs an ADMWD-based encoder, which combines the absolute deviation from the mean (ADM) and the weighted differential (WD) distance, to embed the raw MTS as a multi-channel time series matrix.

\subsubsection{Frequency-Domain-Based Methods}
\label{subsubsec:mtad-spatiotemporal-dependence-frequency-domain-based-methods}

Some approaches in anomaly detection explore the utilization of frequency domain information, complementing the techniques inspired by computer vision processing.
These methods leverage Fast-Fourier Transform (FFT) to extract a frequency vector for each metric in MTS.

RANSynCoders~\cite{DBLP:conf/kdd/AbdulaalLL21} uses spectral analysis on the feature representation to capture frequency domain information in MTS.
This enables learning phase shifts among time series, synchronizing the raw multivariate data and enhancing spatial colinearity even with asynchronous signals.
It specializes in point-anomaly detection, eliminating the need for posterior threshold selection and making it suitable for detecting various anomaly types effectively.

Similarly, TFAD~\cite{DBLP:conf/cikm/ZhangZWS22} includes a time series decomposition module that utilizes a simple convolutional neural network.
The process starts with an augmentation module handling both normal and abnormal data augmentation on the raw time series.
Then, the decomposition module transforms the augmented time series into representations of the time and frequency domains. The anomaly score is calculated by combining the results from the two branches.

\subsubsection{Graph-Based Methods}
\label{subsubsec:mtad-spatiotemporal-dependence-graph-based-methods}

In the context of scalable cloud computing systems, there is a need for an approach that can effectively handle variable-length information and capture spatial dependence efficiently.
MSCRED, as mentioned earlier, suffers from high computational complexity and lacks focus on capturing topological information between components.
Some approaches utilize graph-based models to address these challenges.
Graph-based models can be represented by the equation:
\begin{equation}
    x_{j}=\sum_{i\in\xi\left(j\right)}{a_{ij}x_{i}},\mathrm{where} \sum_{i\in\xi\left(j\right)}{a_{ij}}=1
\end{equation}
where $a_{ij}$ is a normalized weight and $\xi\left(j\right)$ denotes the set of adjacent nodes of node $j$.
Typically, the MTS is processed by a graph neural network (GNN), which projects it into an embedding vector for downstream classifiers.

Many graph-based MTS anomaly detection methods have been proposed in the AIOps domain.
MARINA~\cite{DBLP:conf/cikm/XieCHL022} addresses concept drift and spatiotemporal correlation learning using an attention module.
It uses cascaded MLP for temporal correlation and a GNN with a graph attention network (GAT) for spatial correlation.
StackVAE-G~\cite{DBLP:journals/aiopen/LiHCCF22} uses stacking-VAE with GNN and a graph learning module for MTS data.
GReLeN~\cite{DBLP:conf/ijcai/ZhangZT22} utilizes a VAE structure to learn a probabilistic relation graph for multiple metrics, where the latent variable of the VAE captures the dependence relationship between metrics.
MTAD-TF~\cite{DBLP:journals/complexity/HeZZW20} employs CNN for temporal patterns and GAT for spatial correlations. Then, it combines spatiotemporal features with a gated RNN.
Other methods like MTAD-GAT~\cite{DBLP:conf/icdm/ZhaoWDHCTXBTZ20} and GTA~\cite{DBLP:journals/iotj/ChenCZYC22} also benefit from the noise resistance of convolutions due to weight sharing and shift-invariance.
HAD-MDGAT~\cite{DBLP:journals/access/ZhouZL22} uses GAT with MDA, addressing overfitting with GAN architecture.
TopoMAD~\cite{DBLP:journals/tnn/HeCLWYCLZ23} combines GNNs and LSTM cells for cloud system anomaly detection.
DVGCRN~\cite{DBLP:conf/icml/ChenTCDDZ22} captures spatiotemporal features with hierarchical latent representations and employs a stacked graph convolutional recurrent network.
GDN~\cite{DBLP:conf/aaai/DengH21} exploits temporal and spatial dependencies using graph-based models.
It also leverages a graph deviation scoring component to identify and explain deviations from the learned metric relationships in the graph.
In this case, GDN computes the similarity of each node and obtains the graph structure information by selecting the top $k$ nodes, which are then fed into graph models.
FuSAGNet~\cite{DBLP:conf/kdd/HanW22} introduces a sparse autoencoder to handle the issue of a sparse adjacency matrix while capturing temporal and spatial dependencies for anomaly detection.

In a recent work by \citet{DBLP:journals/tim/ZhuRLJ23}, a GAT is employed to capture spatial dependence.
The method utilizes both an LSTM prediction model and a VAE reconstruction model simultaneously to compute the anomaly score.
This joint usage enables the model to effectively capture short-term and long-term temporal dependencies.
Another recent approach, DCFF-MTAD~\cite{DBLP:journals/sensors/XuYGH23}, introduces a dual-channel feature extraction module.
One channel focuses on spatial features utilizing spatial short-time Fourier transform (STFT), while the other channel leverages a time-based GAT to capture temporal features.
Similarly, DCFF-MTAD also incorporates prediction and reconstruction models to calculate the anomaly score.
By combining spatial and temporal representations learned from both channels, the model achieves improved performance in anomaly detection.

We can now summarize how these methods construct adjacency matrices.
The adjacency matrix captures graph structure information, representing correlations between nodes and reducing computational complexity by eliminating redundant message passing.
Some graph-based methods incorporate specific graph structure learning modules to automatically learn the adjacency matrices~\cite{DBLP:journals/aiopen/LiHCCF22, DBLP:conf/ijcai/ZhangZT22, DBLP:conf/kdd/HanW22, DBLP:journals/access/ZhouZL22, DBLP:conf/icml/ChenTCDDZ22, DBLP:conf/aaai/DengH21, DBLP:journals/tim/ZhuRLJ23}, while others do not explicitly emphasize this aspect~\cite{DBLP:journals/complexity/HeZZW20, DBLP:conf/icdm/ZhaoWDHCTXBTZ20, DBLP:journals/tnn/HeCLWYCLZ23, DBLP:journals/iotj/ChenCZYC22, DBLP:journals/sensors/XuYGH23}.
The availability of structure information is crucial as input for a graph-based model, but in many cases, the correlations between nodes are initially unknown.
Therefore, graph structure learning plays a significant role in addressing this issue.

\subsubsection{Representation-Oriented Methods}
\label{subsubsec:mtad-spatiotemporal-dependence-representation-oriented-methods}

Several MTS anomaly detection approaches concentrate on the latent space to acquire a robust representation.
These methods aim to explicitly capture corresponding characteristics by designing novel representations of hidden features.

InterFusion~\cite{DBLP:conf/kdd/LiZHSJWP21} addresses the challenge of modeling inter-metric and temporal dependence in MTS by introducing a two-view embedding.
This embedding derives latent variables that characterize the inter-metric and temporal dependence.
InterFusion also incorporates a novel prefiltering strategy to address potential anomalies in the training data.
SDFVAE~\cite{DBLP:conf/www/DaiLLJLXZ21} specifically focuses on learning representations for both time-varying and time-invariant characteristics hidden in MTS.
It proposes a novel representation model that factorizes the latent space into two separate variables: static and dynamic.
These variables correspond to the time-varying and time-invariant characteristics of MTS, respectively.
Then it employs a bidirectional LSTM model to capture the static latent representations and a recurrent VAE to learn the dynamic latent representations.

Similar to DOMI~\cite{DBLP:journals/tc/SuZSZWZLLTWP22}, SGmVRNN~\cite{DBLP:conf/infocom/DaiCLALLWXC22} also integrates Gaussian mixture components in a recurrent VAE.
However, SGmVRNN distinguishes itself by dividing the latent variables into two parts: categorical latent variables and dynamic latent variables.
These variables collaborate with the Gaussian mixture component to learn MTS's complex structural and dynamic characteristics.

\subsubsection{RNN-Based Methods}
\label{subsubsec:mtad-spatiotemporal-dependence-rnn-based-methods}

Many methods primarily focus on system metrics or service metrics.
In contrast, SCWarn~\cite{DBLP:conf/sigsoft/Zhao0YWLQXZSP21} introduces a multi-modal learning approach to detect anomalies using heterogeneous multi-source data.
In SCWarn, inter-correlations among the multi-source data are encoded, and the temporal dependence in each time series is learned using LSTM.
In order to enhance the accuracy of anomaly detection, SCWarn employs multi-modal learning to simultaneously capture different levels of information. This is achieved through a fully connected layer, where the input is the product of upstream LSTM models.

\subsubsection{Self-Supervised Methods}
\label{subsubsec:mtad-spatiotemporal-dependence-self-supervised-methods}

We have discussed various anomaly detection methods for MTS in the AIOps domain.
However, none of these methods have utilized out-of-band information to aid in anomaly detection.
ACVAE~\cite{DBLP:journals/jsac/LiZGZWCJVSP22} is the first approach to use out-of-band information for situation-aware anomaly detection.
It leverages background and feedback data to determine anomalies in memory usage, considering the code deployment status.
For instance, during a software change, it can easily recognize a spike as a normal pattern by utilizing the background information.
To achieve this, ACVAE integrates a query model that selects valuable instances to obtain feedback.

ACVAE, along with CAE-AD and other methods published after 2022, focuses on self-supervised learning and addresses the issue of label scarcity.
These methods demonstrate the growing attention towards active learning and contrastive learning in MTS anomaly detection.

\subsection{Summary}

MTS anomaly detection methods in the AIOps domain commonly use specific models to generate two embeddings along the spatial (inter-metric) or temporal dimension.
RNNs are widely used for mapping time series into feature representations, but due to computational complexity and error accumulation, some approaches explore more efficient CNNs for sequence projection.
Additionally, the increasing popularity of GNN has led to their use in learning inter-metric topological connections.
There is also a growing interest in contrastive learning and active learning, which can be attributed to the scarcity of labeled data.

Table~\ref{tab:mtad-summary} compares the methods discussed above, considering various perspectives.
These methods are classified into three groups: prediction-based (P), reconstruction-based (R), and hybrid methods (P/R).
Their typical usage scenarios are also taken into account for categorization purposes.
Table~\ref{tab:utad-summary} also presents their primary technique, namely temporal dependence only (T) or spatiotemporal dependence (T/S), the evaluation datasets used, and the evaluation metrics employed.
Additionally, Table~\ref{tab:mtad-datasets} provides a compilation of commonly utilized evaluation datasets and their corresponding links.
For a detailed understanding of the evaluation metrics, please refer to \S~\ref{sec:evaluation-metrics}.

\begin{table}
    \centering
    \tiny
    \begin{threeparttable}
        \caption{
            Summary of MTS Anomaly Detection Methods in the AIOps Domain
        }
        \label{tab:mtad-summary}
        \begin{tabular}{lcccccc}
            \toprule
            Method                                                 & Scenario                  & Type & Dimension & Techniques  & Evaluation Datasets     & Evaluation Metrics\tnote{*}        \\
            \midrule
            TopoMAD ~\cite{DBLP:journals/tnn/HeCLWYCLZ23}          & \multirow{10}{*}{Cloud}   & R    & T/S       & Graph-based & (k)/(l)                 & $F_{1}$/mAP/ROC                    \\
            SDFVAE~\cite{DBLP:conf/www/DaiLLJLXZ21}                &                           & R    & T         & \textemdash & (a)                     & $F_{1}$/PR                         \\
            JumpStarter~\cite{DBLP:conf/usenix/MaZ0XLLNZWP21}      &                           & R    & T         & \textemdash & (a)/(b)/(c)             & Point-Adjusted $F_{1}$             \\
            SCWarn~\cite{DBLP:conf/sigsoft/Zhao0YWLQXZSP21}        &                           & R    & T/S       & ML-based    & \textemdash             & $F_{1}$/MTTD                       \\
            DOMI~\cite{DBLP:journals/tc/SuZSZWZLLTWP22}            &                           & R    & T         & \textemdash & (h)                     & $F_{1}$/ROC                        \\
            SLA-VAE~\cite{DBLP:conf/www/HuangCL22}                 &                           & R    & T/S       & CNN-based   & \textemdash             & $F_{1}$                            \\
            SGmVRNN~\cite{DBLP:conf/infocom/DaiCLALLWXC22}         &                           & R    & T/S       & ML-based    & (a)                     & $F_{1}$                            \\
            LR-SemiVAE~\cite{DBLP:journals/apin/ChenTDHG23}        &                           & R    & T         & \textemdash & (g)                     & $F_{1}$                            \\
            Active-MTSAD~\cite{DBLP:conf/dsn/WangCXH22}            &                           & R    & T         & \textemdash & (a)                     & Point-Adjusted $F_{1}$             \\
            ACVAE~\cite{DBLP:journals/jsac/LiZGZWCJVSP22}          &                           & R    & T/S       & ML-based    & (m)                     & $F_{1}$                            \\
            \midrule
            MTAD-TF~\cite{DBLP:journals/complexity/HeZZW20}        & \multirow{17}{*}{General} & P    & T/S       & Graph-based & (a)/(b)/(c)             & $F_{1}$                            \\
            CTF~\cite{DBLP:conf/infocom/SunSZCLPWZLT21}            &                           & R    & T         & \textemdash & (j)                     & $F_{1}$                            \\
            MTS-DCGAN~\cite{DBLP:journals/ijon/LiangSWGLL21}       &                           & R    & T/S       & CNN-based   & \textemdash             & $F_{1}$                            \\
            USAD~\cite{DBLP:conf/kdd/AudibertMGMZ20}               &                           & R    & T         & \textemdash & (a)/(b)/(c)/(d)/(e)     & Point-Adjusted $F_{1}$             \\
            MSCRED~\cite{DBLP:conf/aaai/ZhangSCFLCNZCC19}          &                           & R    & T/S       & RNN-based   & \textemdash             & $F_{1}$                            \\
            UNLEARN~\cite{DBLP:conf/ccs/DuCLOS19}                  &                           & R    & T         & \textemdash & (g)                     & $F_{1}$                            \\
            OmniAnomaly~\cite{DBLP:conf/kdd/SuZNLSP19}             &                           & R    & T         & \textemdash & (a)/(b)/(c)             & Point-Adjusted $F_{1}$             \\
            RANSynCoders~\cite{DBLP:conf/kdd/AbdulaalLL21}         &                           & R    & T/S       & ML-based    & \textemdash             & Point-Adjusted $F_{1}$             \\
            StackVAE-G~\cite{DBLP:journals/aiopen/LiHCCF22}        &                           & R    & T/S       & Graph-based & (a)/(b)/(c)             & Point-Adjusted $F_{1}$             \\
            InterFusion~\cite{DBLP:conf/kdd/LiZHSJWP21}            &                           & R    & T/S       & RNN-based   & (a)/(b)/(c)/(d)/(e)/(i) & Point-Adjusted $F_{1}$/TPR/FPR/ROC \\
            CAE-ensemble~\cite{DBLP:journals/pvldb/CamposKGHZYJ21} &                           & R    & T         & \textemdash & (a)/(b)/(c)/(d)/(f)     & $F_{1}$/PR/ROC                     \\
            MTAD-GAT~\cite{DBLP:conf/icdm/ZhaoWDHCTXBTZ20}         &                           & P    & T/S       & Graph-based & (b)/(c)                 & $F_{1}$                            \\
            DAEMON~\cite{DBLP:conf/icde/ChenDHZZZZ21}              &                           & R    & T         & \textemdash & (a)/(b)/(c)/(e)         & Point-Adjusted $F_{1}$             \\
            DVGCRN~\cite{DBLP:conf/icml/ChenTCDDZ22}               &                           & P/R  & T/S       & Graph-based & (a)/(b)/(c)             & $F_{1}$                            \\
            GReLeN~\cite{DBLP:conf/ijcai/ZhangZT22}                &                           & R    & T/S       & Graph-based & (a)/(b)/(c)             & Point-Adjusted $F_{1}$             \\
            DCFF-MTAD~\cite{DBLP:journals/sensors/XuYGH23}         &                           & P/R  & T/S       & Graph-based & (a)/(b)/(c)             & $F_{1}$                            \\
            \midrule
            ADTCD~\cite{xu2023adtcd}                               & \multirow{8}{*}{IoT}      & R    & T         & ML-based    & (d)/(e)                 & ROC                                \\
            HAD-MDGAT~\cite{DBLP:journals/access/ZhouZL22}         &                           & P/R  & T/S       & Graph-based & (a)/(b)/(c)/(d)/(e)     & $F_{1}$                            \\
            GTA~\cite{DBLP:journals/iotj/ChenCZYC22}               &                           & R    & T/S       & Graph-based & (b)/(c)/(d)/(e)         & Point-Adjusted $F_{1}$             \\
            TSMAE~\cite{gao2022tsmae}                              &                           & R    & T         & \textemdash & (f)                     & TPR/FPR/ROC                        \\
            PBATPC~\cite{DBLP:journals/iotj/HanLPLC22}             &                           & P    & T/S       & CNN-based   & \textemdash             & MAPE                               \\
            CAE-AD~\cite{DBLP:journals/isci/ZhouYZWY22}            &                           & R    & T         & \textemdash & (a)/(b)/(c)             & $F_{1}$                            \\
            MARINA~\cite{DBLP:conf/cikm/XieCHL022}                 &                           & P    & T/S       & Graph-based & (p)                     & $F_{1}$/MSE/MAE                    \\
            TFAD~\citet{DBLP:conf/cikm/ZhangZWS22}                 &                           & P    & T         & \textemdash & (b)/(c)                 & Point-Adjusted $F_{1}$             \\
            \midrule
            GDN~\cite{DBLP:conf/aaai/DengH21}                      & \multirow{4}{*}{CPS}      & P    & T/S       & Graph-based & (d)/(e)                 & Point-Adjusted $F_{1}$             \\
            MAD-GAN~\cite{DBLP:conf/icann/LiCJSGN19}               &                           & P    & T         & \textemdash & (d)/(e)                 & $F_{1}$                            \\
            FuSAGNet~\cite{DBLP:conf/kdd/HanW22}                   &                           & P/R  & T/S       & Graph-based & (d)/(e)/(o)             & $F_{1}$                            \\
            \citet{DBLP:journals/tim/ZhuRLJ23}                     &                           & P/R  & T/S       & Graph-based & \textemdash             & Point-Adjusted $F_{1}$             \\
            \bottomrule
        \end{tabular}
        \begin{tablenotes}
            \item [*] mAP: Mean Average Precision; ROC: Receiver Operating Characteristic; PR: Precision-Recall; MTTD: Mean Time to Detect; TPR: True Positive Rate; FPR: False Positive Rate; MAPE: Mean Absolute Percentage Error; MSE: Mean Squared Error; MAE: Mean Absolute Error.
        \end{tablenotes}
    \end{threeparttable}
\end{table}

\begin{table}
    \centering
    \tiny
    \caption{
        Public MTS Anomaly Detection Datasets in the AIOps Domain Sorted by Frequency of Usage
    }
    \label{tab:mtad-datasets}
    \begin{tabular}{ccl}
        \toprule
        No. & Name         & Link                                                                \\
        \midrule
        (a) & SMD          & \url{https://github.com/NetManAIOps/OmniAnomaly}                    \\
        (b) & SMAP         & \url{https://github.com/khundman/telemanom}                         \\
        (c) & MSL          & \url{https://github.com/khundman/telemanom}                         \\
        (d) & WADI         & \url{https://itrust.sutd.edu.sg/itrust-labs_datasets/dataset_info/} \\
        (e) & SWaT         & \url{https://itrust.sutd.edu.sg/itrust-labs_datasets/dataset_info/} \\
        (f) & ECG          & \url{https://www.cs.ucr.edu/~eamonn/time_series_data_2018/}         \\
        (g) & Yahoo        & \url{https://webscope.sandbox.yahoo.com/}                           \\
        (h) & DOMI dataset & \url{https://github.com/NetManAIOps/DOMI_dataset}                   \\
        (i) & ASD          & \url{https://github.com/zhhlee/InterFusion}                         \\
        (j) & CTF dataset  & \url{https://github.com/NetManAIOps/CTF_data}                       \\
        (k) & MMS          & \url{https://github.com/QAZASDEDC/TopoMAD}                          \\
        (l) & MBD          & \url{https://github.com/QAZASDEDC/TopoMAD}                          \\
        (m) & SKAB         & \url{https://github.com/waico/SKAB}                                 \\
        (n) & NAB          & \url{https://github.com/numenta/NAB}                                \\
        (o) & HAI          & \url{https://github.com/icsdataset/hai}                             \\
        (p) & ETDataset    & \url{https://github.com/zhouhaoyi/ETDataset}                        \\
        \bottomrule
    \end{tabular}
\end{table}

Table~\ref{tab:mtad-repos} presents the public code repositories of the MTS anomaly detection method.
Most of them maintain an official code repository, while some are unofficial or not publicly available.
Those MTS anomaly detection methods whose codes are not publicly available are absent from the table.

\begin{table}
    \centering
    \tiny
    \caption{
        Source Code of UTS Anomaly Detection Datasets in the AIOps Domain
    }
    \label{tab:mtad-repos}
    \begin{tabular}{cccc}
        \toprule
        Name         & Language & Method                                     & Code                                             \\
        \midrule
        TopoMAD      & Python   & \citet{DBLP:journals/tnn/HeCLWYCLZ23}      & \url{https://github.com/QAZASDEDC/TopoMAD}       \\
        SDFVAE       & Python   & \citet{DBLP:conf/www/DaiLLJLXZ21}          & \url{https://github.com/dlagul/SDFVAE}           \\
        JumpStarter  & Python   & \citet{DBLP:conf/usenix/MaZ0XLLNZWP21}     & \url{https://github.com/NetManAIOps/JumpStarter} \\
        SCWarn       & Python   & \citet{DBLP:conf/sigsoft/Zhao0YWLQXZSP21}  & \url{https://github.com/FSEwork/SCWarn}          \\
        DOMI         & Python   & \citet{DBLP:journals/tc/SuZSZWZLLTWP22}    & \url{https://github.com/NetManAIOps/DOMI\_code}  \\
        SGmVRNN      & Python   & \citet{DBLP:conf/infocom/DaiCLALLWXC22}    & \url{https://github.com/dlagul/SGmVRNN}          \\
        USAD         & Python   & \citet{DBLP:conf/kdd/AudibertMGMZ20}       & \url{https://github.com/manigalati/usad}         \\
        MSCRED       & Python   & \citet{DBLP:conf/aaai/ZhangSCFLCNZCC19}    & \url{https://github.com/NetManAIOps/OmniAnomaly} \\
        OmniAnomaly  & Python   & \citet{DBLP:conf/kdd/SuZNLSP19}            & \url{https://github.com/NetManAIOps/OmniAnomaly} \\
        RANSynCoders & Python   & \citet{DBLP:conf/kdd/AbdulaalLL21}         & \url{https://github.com/eBay/RANSynCoders}       \\
        InterFusion  & Python   & \citet{DBLP:conf/kdd/LiZHSJWP21}           & \url{https://github.com/zhhlee/InterFusion}      \\
        CAE-ensemble & Python   & \citet{DBLP:journals/pvldb/CamposKGHZYJ21} & \url{https://github.com/d-gcc/CAE-Ensemble}      \\
        MTAD-GAT     & Python   & \citet{DBLP:conf/icdm/ZhaoWDHCTXBTZ20}     & \url{https://github.com/ML4ITS/mtad-gat-pytorch} \\
        MAD-GAN      & Python   & \citet{DBLP:conf/icann/LiCJSGN19}          & \url{https://github.com/LiDan456/MAD-GANs}       \\
        \bottomrule
    \end{tabular}
\end{table}

\section{Evaluation Metrics}
\label{sec:evaluation-metrics}
Many time series anomaly detection methods validate their models using metrics like $F_{1}$-score and Area Under Curve (AUC).
$F_{1}$-score is the harmonic mean of Precision and Recall, whose calculation requires True Positives (TP), False Positives (FP), and False Negatives (FN), which is given by:
\begin{equation}
    \begin{aligned}
         & \mathrm{Precision}=\frac{\mathrm{TP}}{\mathrm{TP}+\mathrm{FP}}                                                 \\
         & \mathrm{Recall}=\frac{\mathrm{TP}}{\mathrm{TP}+\mathrm{FN}}                                                    \\
         & F_{1}\mathrm{-score}=2\times\frac{\mathrm{Precision}\times\mathrm{Recall}}{\mathrm{Precision}+\mathrm{Recall}}
    \end{aligned}
\end{equation}
``Precision'' is a metric that quantifies the number of correct positive predictions made.
``Recall'' is a metric that quantifies the number of correct positive predictions made out of all positive predictions that could have been made.
Both the Precision and the Recall are focused on the positive class (anomaly points in the case of anomaly detection) and are unconcerned with the true negatives (normal points).
Using Precision and Recall, $F_{1}$-score makes it possible to assess the performance of a classifier on the minority class.

As for AUC, two types of them are often used by related works: Area Under Precision-Recall Curve (PR-AUC) and Area Under Receiver Operating Characteristic Curve (ROC-AUC).
A Precision-Recall curve (or PR curve) plots the Precision and the Recall for different probability thresholds.
A model with perfect skill is depicted as a point at a coordinate of $\left(1,1\right)$.
A skillful model is represented by a curve that bows towards a coordinate of $\left(1,1\right)$.
A no-skill classifier will be a horizontal line on the plot with a precision that is proportional to the number of positive examples in the dataset.
The focus of the PR curve on the minority class makes it an effective diagnostic for imbalanced binary classification models like anomaly detection methods.
The ROC curve (or receiver operating characteristic curve) is also a plot that can summarize the performance of a binary classification model on the positive class.
ROC uses the False Positive Rate (FPR), and the True Positive Rate (TPR), which can be calculated by:
\begin{equation}
    \begin{aligned}
         & \mathrm{FPR}=\frac{\mathrm{FP}}{\mathrm{FP}+\mathrm{TN}} \\
         & \mathrm{TPR}=\frac{\mathrm{TP}}{\mathrm{TP}+\mathrm{FN}}
    \end{aligned}
\end{equation}
The true positive rate is also called the sensitivity or the Recall.
We can think of the ROC as the fraction of correct predictions for the positive class versus the fraction of errors for the negative class.
Ideally, we want the fraction of correct positive class predictions to be 1 and the fraction of incorrect negative class predictions to be 0.
A trade-off exists between the TPR and FPR, such that changing the classification threshold will change the balance of predictions towards improving the TPR at the expense of FPR, or the reverse case.
It is a popular diagnostic tool for classifiers on balanced and imbalanced binary prediction problems because it is not biased toward the majority or minority class.

Although both PR curve and ROC are helpful diagnostic tools, comparing two or more classifiers based only on their curves can be challenging.
Instead, the area under the curve can be calculated to give a single score for a classifier model across all threshold values.
This is called AUC.
The score is a value between 0.0 and 1.0.
This single score can be used to compare anomaly detection models directly.
PR-AUC is recommended for highly skewed domains where ROC-AUC may provide an excessively optimistic view of the performance.

Since classical point-based Precision and Recall are inadequate to represent domain-specific performance for anomaly detection
on time series data in the AIOps domain (\eg metric data in this study)~\cite{DBLP:conf/www/XuCZLBLLZPFCWQ18}, two strategies, namely the point-adjusted strategy, and the range-based strategy have been proposed.
For the point-adjusted strategy, if any point in a ground-truth anomaly segment is correctly identified as an anomaly by a model, it will consider all other points in the segment as true positives when evaluating Precision and Recall.
This strategy has been widely utilized in previous work~\cite{DBLP:conf/www/XuCZLBLLZPFCWQ18, DBLP:conf/ipccc/LiCP18, DBLP:journals/jsac/ZhangZLFSZZPSLY22}.
By comparison, the range-based strategy~\cite{DBLP:conf/nips/TatbulLZAG18} is stricter and considers more aspects of the predictions, including the existence, size, position, and cardinality.
Both strategies are tailored to the AIOps domain and are more suitable for evaluating anomaly detection methods on time series data.

\section{Concluding Remarks and Future Work}
\label{sec:concluding-remarks-and-future-work}
A comprehensive survey on time series anomaly detection in the AIOps domain aims to provide readers with a clear understanding of the reasons behind using specific anomaly detection techniques while also offering a comparative analysis of various approaches.
However, previous research has been conducted unstructured, lacking a comprehensive taxonomy for this problem.
Consequently, providing a theoretical understanding of the anomaly detection problem becomes challenging.
In this survey, we have explored different aspects related to the issue of time series anomaly detection in the AIOps domain and provided an overview of the extensive literature on various techniques.
For univariate and multivariate time series, we have classified each anomaly detection technique based on its fundamental methods and identified their primary contributions in addressing the problem.
These assumptions can serve as guidelines when applying a particular technique to a real-world use case, enabling an assessment of its effectiveness in that specific scenario.

However, the existing research does not explicitly indicate which anomaly detection techniques should be employed when dealing with time series data exhibiting diverse anomaly types and statistical or visual patterns of normalcy.
An avenue for future investigation would involve thoroughly exploring the characteristics of different techniques and evaluating their performance in relation to various data attributes.
\citet{DBLP:journals/pvldb/SchmidlWP22} has taken a partial step in this direction, where the authors compiled and re-implemented 71 anomaly detection algorithms from different domains, subsequently evaluating them on 976 time series datasets.
The evaluation encompassed multiple factors, including effectiveness, efficiency, and robustness, yielding a concise overview of the techniques and their commonalities.
Although these experimental results can facilitate the selection of appropriate algorithms, the evaluated anomaly detection techniques are not specifically tailored to the domain of AIOps, nor do they offer a comprehensive classification of different anomaly types.

The AIOps domain offers numerous promising research directions for time series anomaly detection.
Contextual and collective anomaly detection techniques are increasingly finding applicability across various domains, presenting ample opportunities for the development of novel techniques~\cite{DBLP:journals/csur/ChandolaBK09}.
The presence of data distributed across different servers or service instances has prompted the need for distributed anomaly detection techniques.
While such techniques process information available at multiple sites, they must simultaneously safeguard the information at each site, necessitating privacy-preserving anomaly detection methods.
With the rapid advancement of microservice architecture, efficient anomaly detection for numerous entities has become imperative.
Many techniques discussed in this survey require dedicated trained models for each monitoring entity, resulting in excessive time and computational resource requirements.
Recent advancements propose techniques to reduce the training or initialization time of anomaly detection in AIOps, as well as endeavors to share knowledge among models or employ a single pre-trained model for detecting anomalies across all entities.
Another emerging research area in multivariate time series anomaly detection involves automatically learning the topological structure of the system (graph learning), thereby enhancing robustness and adaptability to changes in the relationships between monitoring entities or adding new entities for detection.


\bibliographystyle{ACM-Reference-Format}
\bibliography{main}

\end{document}